\documentclass[]{fairmeta}
\usepackage{amssymb,amsmath}
\usepackage{amsthm}
\usepackage{hyperref}
\usepackage{graphicx} 
\usepackage{url}
\usepackage{amsthm}
\usepackage{thmtools}
\usepackage{mathtools}
\usepackage{upgreek}
\usepackage{dsfont}
\usepackage{cleveref}
\usepackage{bbm}
\usepackage{booktabs}
\usepackage{multirow}
\usepackage{array}
\usepackage{makecell}
\usepackage{tcolorbox}
\usepackage{footnote}
\usepackage{algorithm2e}
\usepackage{algorithmic}
\usepackage{etoc}
\usepackage[rightcaption]{sidecap} %

\usepackage{enumitem}
\setitemize{noitemsep,topsep=0pt,parsep=0pt,partopsep=0pt}

\newcommand{\xxcomment}[4]{\textcolor{#1}{[$^{\textsc{#2}}_{\textsc{#3}}$ #4]}}
\newcommand{\julia}[1]{\xxcomment{purple}{J}{K}{#1}}

\newcommand{\ignore}[1]{}

\def\E{{\mathbb{E}}} %
\def\X{{\mathcal{X}}} %

\newcommand{\algoname}{\texttt{SOAR}\xspace} 

\newtheorem{thm}{Theorem}

\newtheorem{prop}[thm]{Proposition}

\newcommand{\hardonly}{\textit{Hard-Only}\xspace}
\newcommand{\intrinsic}{\textit{Intrinsic-T}\xspace}
\newcommand{\base}{\textit{Base-T}\xspace}
\newcommand{\grounded}{\textit{Grounded-T}\xspace}
\newcommand{\groundednp}{\textit{Grounded-T (no promotion)}\xspace}

\usepackage{natbib}

\usepackage{tcolorbox}

\newtcolorbox{takeaway}{
  colback=gray!5,
  colframe=gray!50,
  boxrule=1pt,
  left=10pt,
  right=10pt,
  top=8pt,
  bottom=8pt,
  arc=3pt
}

\title{Teaching Models to Teach Themselves:\\ Reasoning at the Edge of Learnability}

\author[1,*]{Shobhita Sundaram}
\author[2]{John Quan}
\author[2]{Ariel Kwiatkowski}
\author[2]{Kartik Ahuja}
\author[2]{Yann Ollivier}
\author[2,3]{Julia Kempe}

\affiliation[1]{MIT}
\affiliation[2]{Meta FAIR}
\affiliation[3]{New York University}

\contribution[*]{Work done during an internship at Meta}

\newcommand{\confonly}[1]{}     %
\newif\ifarxiv  \arxivtrue
\newif\ifconference \conferencefalse

\abstract{

RL methods for scaling large reasoning models stall on datasets with low initial success rates, and thus little training signal. We investigate a fundamental question: \textit{Can a pretrained LLM leverage latent knowledge to generate an automated curriculum  for problems it cannot solve? }
We explore this with \algoname: An asymmetric self-play framework that uses meta-RL to surface these pedagogical signals. 
A teacher model proposes synthetic problems for a student model, and is rewarded with its improvement on a subset of hard problems, thus grounding the curriculum in real student progress rather than intrinsic proxy rewards.
Our study on the hardest subsets of math benchmarks (0/128 success) reveals three core findings. 
First, it is possible to realize bilevel meta-RL that unlocks learning under sparse, binary rewards by sharpening a latent capacity of pretrained models to generate useful problems.
Second, grounded rewards outperform intrinsic learnability rewards used in prior LLM self-play, reliably avoiding typical instability and diversity collapse modes. 
Third, the structure and well-posedness of questions are more critical for learning progress than solution correctness. Our results suggest that the ability to generate useful stepping stones does not require the preexisting ability to solve 
the hard problems, paving a principled path to escape reasoning 
plateaus without additional curated data.

\ignore{Our study study on the hardest subsets of math benchmarks reveals three core findings: First, it is possible to pull off this bilevel  meta-RL with grounded rewards - we get it to perform and unlock learning in the hardest setting with sparse binary rewarwds. we show how our meta-RL resurfaces a latent ability of the model to already generate curricular stepping stones. Second, we show that grounded rewards improves majorly over a baseline teacher that isn't trained, and, more crucially, over intrinsic rewards that are used in essentially all other prior LLM self-play works. And third, we find when analyzing the questions that structure and well-posedness matter more than correctness.  }

\ignore{\julia{v1}
Can a model learn to escape its own learning plateau? RL methods for
finetuning large reasoning models stall on datasets with low initial
success rates, and thus little training signal.
We posit that pretrained LLMs themselves can be tapped as generators of learnable reasoning questions to unlock progress on such hard subsets. Inspired by methods from
self-play and curriculum learning, we propose \algoname: A meta-RL
framework that trains an asymmetric teacher and student to produce a
synthetic curriculum that kickstarts learning on hard datasets. In
contrast to %
frameworks that use intrinsic, data-free rewards, our bilevel
optimization loop grounds teacher rewards in measured student
improvement. The teacher generates candidate questions-answer pairs that
the student trains on with RL; the teacher is rewarded using the student
progress on the original hard dataset.  %
On hard subsets of math
benchmarks (0/128 success), \algoname produces synthetic problems that
substantially improve performance over training directly on the hard
data, and over intrinsic-reward baselines.
We also show that the synthetic problems transfer well to other datasets,
and that synthetic questions successfully adapt over time to the
student's abilities.
Our
results suggest that the ability to generate useful stepping stones does
not require the preexisting ability to actually solve the hard problems,
paving a principled path to escape reasoning plateaus without curated additional data. %
}

}

\date{\today}
\correspondence{Shobhita Sundaram at  shobhita@mit.edu}

\begin{document}

\maketitle
\begin{figure*}[h]
    \centering
    \includegraphics[width=\linewidth]{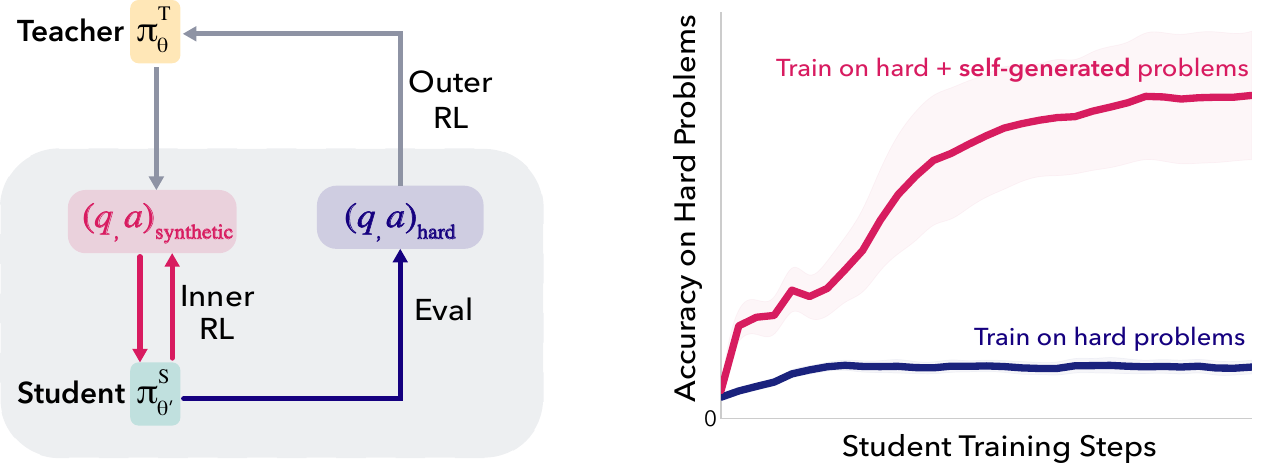}
    \caption{\textbf{Learning on hard problems by self-generating a
    curriculum.} We introduce \textbf{\algoname}: 
    An asymmetric self-play framework that uses meta-RL to improve on difficult datasets where performance plateaus.
    \textbf{(left)} A teacher model generates synthetic
    problems for the student to train on with RL. The teacher is
    rewarded by the student's measured improvement on the real problems, providing a grounding signal. \textbf{(right)}
    Training on problems generated with \algoname (using hard problems as
    a teacher reward signal) outperforms direct training on the hard
    problems, allowing the student to break out of the performance
    plateau.}
    \label{fig:qual}
\end{figure*}

\section{Introduction}

Reinforcement learning with verifiable rewards (RLVR) has
spurred an impressive rise in LLM reasoning capabilities~\citep{deepseek2025r1,kimiteam2025kimi}. However, this paradigm has a key limitation: \textit{the model cannot learn from problems that it cannot
already solve to some extent}. When problems are too difficult, sparse or non-existent rewards leave the model ``stuck".

Past work has shown that curricula strongly affect
generalization in RL
\citep{bengio2009curriculum,Navekar2020curriculum}, with success in selecting ``learnable" problems and adapting easy-to-hard progressions \citep{parashar2025curriculumreinforcementlearningeasy,chen2025sec}. 
Such curricula, however, require careful design and curated intermediate datasets \citep{kordi2025revisiting}. 
Recent work exploits dense signals from test-case pass rates in coding problems \citep{sun2025rlgrokkingrecipe} to address sparse rewards, but relies on curated test-cases.

Asymmetric self-play \citep{silver2018alphazero,sukhbaatar2017asymmetric,openai2021asymmetricselfplay} offers a potential solution to these limitations by enabling self-generated curricula.
Here, we ask: 
\begin{center}
\textit{Can a model break its reasoning plateau by generating its own stepping-stone curriculum?}   
\end{center}

We posit that pretrained LLMs  possess the capacity to directly generate a ``stepping stone curriculum'' to tackle hard problems. To investigate if this pedagogical signal is \textit{present} and \textit{extractable}, we design \algoname: an \emph{asymmetric self-play framework} that uses hard problems as a guiding signal. Both the teacher and student are initialized from the target model; the
teacher proposes questions-answer pairs that the student trains on with
RL, and is rewarded with student improvement on a difficult subset. Critically, rather than using intrinsic rewards common
to self-play, we use the difficult training dataset as a black-box grounding reward signal to guide the teacher towards useful questions for the student.

Intuitively, a pretrained model has already encountered a vast array of easy problems. Consider a difficult calculus question: While the model may be unable to directly answer correctly, it might still possess the latent knowledge to generate easy chain-rule exercises, without requiring a human-in-the-loop to identify and source such questions. We find that by leveraging pretraining knowledge, RL can effectively surface and amplify these latent pedagogical signals to generate useful question-answer pairs. Importantly, we do so without actually showing the model the hard questions; our framework recovers a useful curriculum just by using performance on the hard dataset as a reward signal. 

Empirically, while direct training on the hard dataset fails, \emph{the teacher in our framework learns to generate useful questions
that get the student ``unstuck'' on the hard dataset, without actually seeing the hard problems}. 
Our main contributions, supported by an extensive multi-seed empirical study and ablations (over 600 runs), are the following:
\vspace{-4pt}
\begin{itemize}
  \setlength{\labelsep}{0.4em}
  \setlength{\itemsep}{1pt}
  \setlength{\topsep}{-4pt}
\item \textbf{Decoupled teaching and solving:} A model's ability to generate effective ``stepping stones" for hard problems is distinct from its ability to solve them. Self-play can exploit this asymmetry to generate problems that \textit{expand the learning frontier}, enabling progress where direct RL fails. While the base model has the capacity to propose useful questions, meta-RL sharpens this noisy distribution into a reliable learning signal.

\item \textbf{Proof-of-concept of self-generated curricula} with \algoname (\textbf{S}elf-\textbf{O}ptimization via \textbf{A}symmetric \textbf{R}L), an asymmetric self-play framework that rewards the teacher for student progress on hard problems. 
On hard subsets of MATH and HARP, self-generated problems improve performance (e.g., 4$\times$ pass@1 and 2$\times$ pass@32 on MATH, 2$\times$ pass@1 and 1.5$\times$ pass@32 on HARP). These problems also transfer to unlock learning on datasets that they were not optimized for.
\item \textbf{Mitigating self-play collapse with grounded rewards:} Grounding teacher rewards in student progress on real problems improves performance over intrinsic rewards common in self-play, which are prone to instability and collapse of question diversity. 

\item \textbf{Question structure over solution correctness}: Problem structure and difficulty calibration matter more for escaping plateaus than answer correctness; generated questions provide useful gradient signal even when the majority of answers are incorrect.
\end{itemize}
These results, backed by a comprehensive empirical study, show that grounded meta-RL can escape genuine learning plateaus by letting models discover for themselves what data they need to learn from to expand their learning frontier. %

\section{Related Work}
For an extended review of related literature see \Cref{app:related}:

\paragraph{Self-Play and Teacher-Student Setups.}
Self-play aims to achieve autonomous capability growth, exemplified by game-playing agents such as AlphaZero~\citep{silver2018alphazero}, and asymmetric teacher-student setups that induce automatic curricula~\citep{sukhbaatar2017asymmetric,openai2021asymmetricselfplay}. LLM self-play methods encounter specific challenges: language rewards are extremely sparse and brittle. For mathematical problems, correctness is essentially binary and offers no gradient toward partial solutions. Thus, essentially all modern methods optimize for self-consistency or solution quality. Earlier works
\citep{chen2024spin,wang2025stablellmselfplay,singh2024beyond,ye2024eva} presuppose the existence of well-formed input prompts or curated high-quality questions.

A series of near-contemporary works leverages pretrained LLMs as question generators to create "fully data-free" co-evolving systems 
\citep{zhao2025absolute,huang2025rzero,kuba2025languageselfplay,fang2025serl,chen2025selfquestioning}. These works leverage intrinsic or proxy rewards such as majority vote, learnability,
reward-model preferences, or gradient magnitudes. 
Due to optimizing intrinsic or proxy objectives, they risk drifting to degenerate
or unlearnable tasks, are sensitive to reward hacking and lack progress guarantees \citep{chae2025understandingselfplay}. 
Prolonged RL with self-rewards often causes sudden and complete performance collapse \citep{shafayat2025largereasoningmodelsselftrain,chae2025understandingselfplay}, when rewards vanish or when generator and solver objectives misalign, especially in discrete, symbolic domains with essentially binary correctness signals. This fragility 
raises the broader question of whether self-improvement driven by intrinsic or self-generated rewards can be sustained within RL.
To our knowledge, our work is the first for LLM self-play to ground the curriculum generation in a concrete failure regime instead of internal difficulty proxies. 

\paragraph{Curriculum Learning in RL.}
Automated curriculum design has a long history \citep{bengio2009curriculum,Graves2017automatedcurriculum,Navekar2020curriculum,parashar2025curriculumreinforcementlearningeasy} 
focusing on \emph{reordering} or
\emph{selecting} existing data to enable or accelerate learning, or, in the context of RL, to help agents acquire complex behaviors
by first mastering simpler tasks. For LLM training, curricula are
applied over curated prompts or problem categories, using proxy signals
such as gradient norms or advantage/difficulty estimates to guide selection ~\citep{kimiteam2025kimi,dennis2020paired,wen2025lightr1,yu2025dapo,bae2025onlinedifficultyfilteringreasoning,chen2025sec,jiang2025ADO}.
By contrast, our goal is not to arrange data but to {\em self-generate tasks} to elicit
learning on a fixed, verifiable hard dataset where standard RLVR fails.

Another line of work instead generates a distribution of environments suitable for agent capabilities \citep{dennis2020paired,racaniere2020settersolver,Jiang2020PrioritizedLR,jiang2021ued}. These works find that unconstrained objectives (e.g., minimax adversarial objectives) lead to degenerate curricula, instead optimizing regret or learning potential. We diverge by grounding rewards in student progress on a fixed target set.

\paragraph{Intrinsic Rewards versus Bilevel Optimization.}  
Various intrinsic rewards have been studied across robotics, simulation, and task-agnostic settings for curricula generation and exploration \cite{Schmidhuber1991APF,pathak2017curiosity,pmlr-v97-colas19a,NEURIPS2019_b6f97e6f,colasAutotelic,sancaktarRegularity}.
In self-play, the use of proxy rewards is often not merely a design preference but a pragmatic simplification. It avoids facing an explicit %
bilevel optimization problem: an appealing but  challenging objective where the output of one optimization (in this instance the optimization of the student trained with RLVR on the teacher's question-answer pairs) is fed into another optimization loop (the performance improvement of the student on the hard dataset). 
Such bilevel optimization appears in
meta-learning \citep{Finn17maml,nichol2018firstordermetalearningalgorithms},
 hyperparameter learning
\citep{maclaurin2015hyperopt} and---partially inspiring our work---in dataset distillation, where an outer loop optimizes a small
dataset for an inner training loop to achieve good target
performance \citep{wang2018dataset,
deng2022remember,feng2024embarrassingly}. In general, such approaches become intractable, as the inner loop involves a multi-step computation
with a large number of steps, requiring backpropagation through time
(BPTT),
unrolling the inner loop and taking meta-gradients. Our approach avoids the need to unroll the inner loop thanks to the use of
RLOO in the outer loop,  using the performance improvement of
the student as the reward to reinforce question-answer sets. This is the first
instance of ``double meta-RL loop'' we are aware of in the context of self-play for LLMs.

\section{Method}

Can a pretrained LLM leverage latent knowledge to generate question-answer pairs for problems it cannot solve? And can this be achieved in domains with sparse, binary rewards lacking automatic question verification? To explore this, we introduce \algoname: an asymmetric self-play framework that uses meta-RL to surface such pedagogical signals. Critically, \algoname grounds the teacher reward in measured student progress rather than intrinsic proxy rewards. If the model can generate useful stepping stones despite being unable to solve the original problems, this would suggest that the latent knowledge exists, and is extractable without human curation.

Let $\pi_{\theta}$ be a language model with parameters $\theta$. We
assume access to a dataset $\mathcal{D}=\{(q_i,
a_i)\}^{|\mathcal{D}|}_{i=1}$ of \textit{difficult} question-answer pairs ($\pi_{\theta}$ produces 0/128 successful generations). 
$\mathcal{D}$ is split into train and test sets: $\mathcal{D}_{train}$, $\mathcal{D}_{test}$. To improve the performance of $\pi_{\theta}$ on
$\mathcal{D}_{test}$, the natural approach is to train $\pi_{\theta}$
directly on $\mathcal{D}_{train}$ using RL (\textit{e.g.}, REINFORCE,
GRPO, RLOO, etc). However, for difficult datasets, this may not improve performance due to the sparsity of positive rewards. We instead use this ``failure regime" as a testbed to see if the model can autonomously recover intermediate problems that make these hard problems more learnable.

\subsection{Overview}
Our framework adopts the teacher-student setup of asymmetric self-play, to ``kickstart" learning on datasets where the initial success rate is too low for successful training. We instantiate two copies of the same model: a teacher $\pi^T_{\phi}$ and a student $\pi^S_{\theta}$. At step zero, $\theta = \phi = \theta_{base}$. 

The teacher's role is to generate synthetic problems that provide the student with the necessary gradient signal to escape the performance plateau. Intuitively, while the teacher may be unable to solve a difficult problem directly, it may still possess the knowledge to {\em generate} easier problems that provide a non-zero reward to the student and shift its policy towards progress on the original problem. 

We formulate this problem as a bilevel optimization problem. The objective is to generate a small synthetic dataset $\X = \{(q_i, a_i)\}_{i=1}^n$ of question-answer pairs such that training $\pi_{\theta}^S$ on $\X$ with RL improves performance on the target domain.
\begin{align}
 \max_{\phi} \quad & \E_{\X \sim \pi^T_{\phi}}
 \left[R\left(\pi^S_{\theta'(\X)},
 \mathcal{D}_{train}\right)
 \right]\nonumber%
 \\
\text{subject to} \quad & \theta'(\X) = \textsc{RL-update}(\theta, \X), \label{eq:obj} 
\end{align}\label{eq:bilevel}
where \textsc{RL-update} describes the RL training procedure of the student on $\X$, yielding parameters $\theta'(\X)$, and $R$ denotes the updated student's performance on $\mathcal{D}_{train}$. 

Such bilevel %
objectives have strong precedence in
meta-learning \citep{Finn17maml,nichol2018firstordermetalearningalgorithms}, hyperparameter learning \citep{maclaurin2015hyperopt} and dataset distillation  \citep{wang2018dataset}. In general, such approaches become intractable, requiring ``backpropagation through gradient descent''%
. 
To avoid the associated computational difficulties, we %
instantiate objective (\ref{eq:obj}) 
as a nested meta-RL loop:
 \begin{itemize}
    \item \textbf{Outer RL loop}: Train the \textbf{teacher} with RLOO \citep{ahmadian-etal-2024-back} to generate question-answer pairs. 
    \item \textbf{Inner RL loop}: Train the \textbf{student} with standard RLVR (also RLOO) on teacher-generated problems. The subsequent performance improvement of the student on $\mathcal{D}_{train}$ is the black-box reward signal for the teacher. 
\end{itemize}
We do not assume automatic verification of synthetic question well-posedness or answer correctness (as \textit{e.g.,} in coding tasks in \citet{zhao2025absolute}). Instead, the teacher generates both the question and answer, treating the question utility as an emergent property of the teacher's reward signal.  Critically, we ground the teacher's objective in measured student progress on $\mathcal{D}_{train}$, rather than intrinsic proxies such as learnability, as done in prior work. \algoname only rewards a synthetic question-answer pair if training on it improves student performance on ground-truth problems. This \textit{black-box grounding signal} tethers question generation to real learning progress, implicitly penalizing degenerate problems and reward hacking. The teacher is not shown the hard problems during training, but rather discovers useful stepping stones purely from this student improvement signal.

\begin{figure*}[h!]
    \centering
\includegraphics[width=1.0\linewidth]{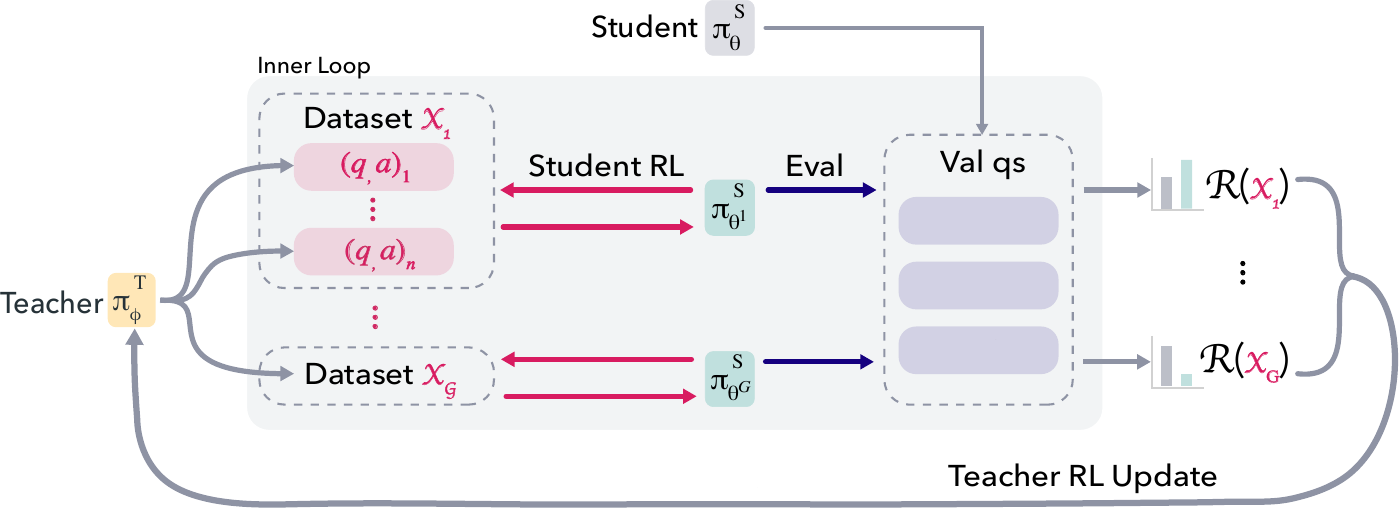}
    \caption{\textbf{The \algoname meta-RL Loop.} The teacher and student are initialized from the same model. In the \textbf{outer RL loop} the teacher generates candidate question-answer pairs that are partitioned into datasets. In the \textbf{inner RL loop}, the student is trained for 10 steps on the candidate problems and evaluated on sampled hard problems. The teacher is rewarded based on the resulting student improvement over the student baseline, grounding the synthetic curriculum in real learning progress.}
    \label{fig:method}
    \vspace{-1pt}
\end{figure*}

See Algorithm \ref{alg:detailed-appendix-ma}, illustrated in Figure \ref{fig:method}, for the full method.

\subsection{Outer Loop: Teacher Training} \label{sec:method-outer}
We train the teacher with RLOO to generate problems that demonstrably improve student performance. Let $g$ denote the RLOO group size and $n$ the size of the generated dataset $\X$. 
At each iteration, we sample $g\cdot n$ rollouts $y_1,\ldots,y_{gn}$ from  $\pi^T_{\phi}$, subdivided into $g$ datasets of $n$ items each: 
$\X_1=\{y_1,\ldots,y_n\},\ldots ,\X_g=\{y_{g(n-1)},\ldots, y_{gn})\}$. Since we cannot automatically verify the answers to proposed problems, we prompt the teacher to generate
both the question \textit{and} answer. Each rollout $y_i$ is parsed into $y_i=(q_i,a_i)$ (described in
Appendix \ref{sec:filtering}; we may need to sample multiple times to obtain a parseable $y_i$). 

Each dataset $\X_k$ receives a reward.
At each outer-loop iteration we subsample a set of \textit{reward
questions} $\mathcal{Q}_R \sim \mathcal{D}_{train}$ from the original training set.
We train the student on each dataset $\X_k$, for a fixed number of steps, resulting in trained student $\pi^S_{\theta'_k}$. 
The dataset-level reward $R(\X_k)$ is then the average greedy success of
$\pi^S_{\theta'_k}$ on $\mathcal{Q}_R$ relative to the
success of a baseline student model $\pi^S_{\theta}$: 
$$
\mathcal{R}(\X_k) = \textsc{Acc}(\pi^S_{\theta'_k}(\mathcal{Q}_R)) - \textsc{Acc}(\pi^S_{\theta}(\mathcal{Q}_R)),
$$

where $\pi^S_\theta$ is the initial student when starting the inner loop. We subtract the initial student accuracy so that teacher rewards are normalized across outer-loop steps, necessary for the student promotion mechanism introduced in Section \ref{sec:method-inner}.

To mitigate reward variance, we average rewards over $r$ parallel student trainings per dataset. This averaged reward is assigned to each rollout in $\X_k$ to update the teacher.

\subsection{Inner Loop: Student Training}
\label{sec:method-inner}

The student $\pi^S_{\theta}$ trains on the teacher-generated dataset $\X_k$ using RLOO for 10 steps (batch size 8), long enough to induce measurable movement while minimizing computational cost. After each inner loop the student reverts to the baseline policy for the next iteration. 

A key question is whether the teacher can adapt to student improvement. To address this, we introduce a
\emph{promotion} mechanism to 
accumulate student improvement and useful questions across inner loops.
We track a rolling moving average of teacher rewards $\bar{R}_t$; when it
exceeds a fixed threshold $\tau$, we update the baseline student $\pi^S_\theta$ to the student trained on the best $\X_k$. Subsequent rewards measure improvement relative to this new baseline (Appendix \ref{app:training-details}). We denote the accumulated datasets that led to student promotions as $\mathcal{D}_{best}$; these constitute the Promotion Questions (PQ) that we evaluate in our experiments.

\section{Experiment Setup}

\subsection{Models and Datasets}

Our experiments mainly use \texttt{Llama-3.2-3B-Instruct}, with ablations extending to \texttt{Llama-3.1-8B-Instruct} in Appendix \ref{app:model-size}. We focus on math reasoning benchmarks to study the prototypical setting of 
sparse, binary rewards, without automatic question-answer verification (as in code):  MATH~\citep{hendrycks2021measuring}, HARP~\citep{yue2024harp}, and OlympiadBench~\citep{he2024olympiadbenchchallengingbenchmarkpromoting}. 
For each dataset, we sample 128 times per problem with the target model, retaining those with a 0/128 success rate. We call these subsets \textit{fail@128} datasets; 128 serves as a practical but stringent threshold at which, empirically, direct training yields only marginal improvement.  Each is randomly split 50-50 into training and test sets. Given the low baseline pass rates on fail@128 problems, this larger test set is necessary to distinguish performance gains from stochastic variance. Details in Appendix \ref{app:datasets}.

\subsection{Teacher-Student Training}
We initialize the teacher and student from the base model and train \algoname on MATH and HARP, keeping OlympiadBench
held-out. 
We allocate 200 outer-loop steps based on compute constraints.
Each outer-loop iteration samples $n=64$ problems ($\mathcal{X}$) from the teacher, and 64 reward questions ($\mathcal{Q}_R$) from the fail@128 train set ($\mathcal{D}_{train}$). 
We promote the student baseline when the 3-step moving average of teacher rewards exceeds $\tau=0.01$. 
Full hyperparameters are in Appendix \ref{app:hparams}, a sensitivity analysis for $\tau$ and $n$ in Appendix \ref{app:ab-teacher-training}, and \algoname training dynamics in Appendix \ref{app:teacher-training-curves}.

\begin{figure*}[t]
    \centering
    \begin{minipage}{0.65\textwidth}
        \centering
        \includegraphics[width=\textwidth]{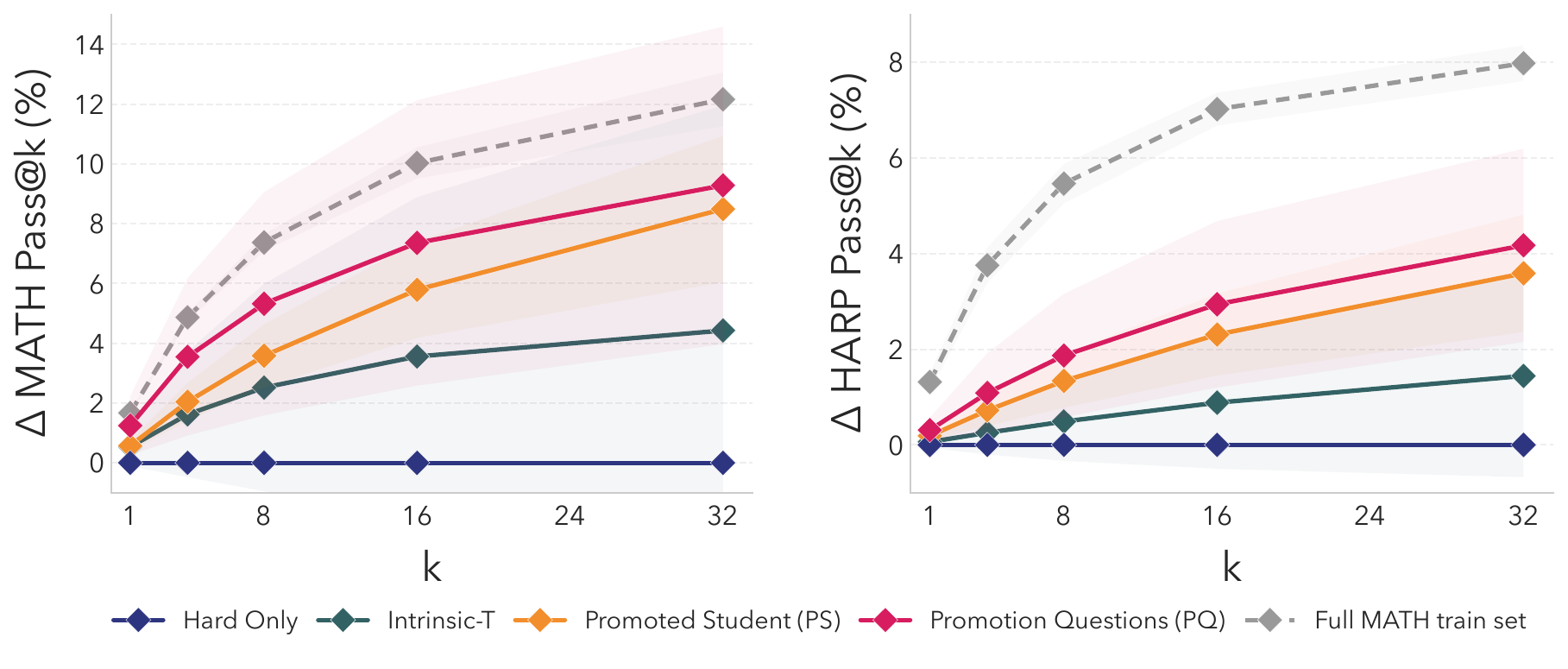}
        \caption{\textbf{Performance on MATH/HARP fail@128 (improvement over \hardonly).} Synthetic problems generated with \algoname (PQ) and inference with the promoted student (PS) outperform direct training on fail@128 train sets (\hardonly), and sampling from teachers trained with intrinsic rewards (\intrinsic). For reference, \hardonly MATH pass@$k$ for $k \in \{1,4,8,16,32\}$ is $\{0.5,  1.7, 3.2, 5.7, 9.6\}$. Full trajectories in Figure \ref{fig:accepted-qs-full}; absolute performance and further evaluations in Tables \ref{tab:app-promotion-results-math}-\ref{tab:app-promotion-results-harp}. Shaded regions are $\pm$ 1 SD over 6-12 seeds (Appendix \ref{app:seeds}).}
        \label{fig:promotion-results}
    \end{minipage}
    \hfill
    \begin{minipage}{0.3\textwidth}
        \centering
        \includegraphics[width=\textwidth]{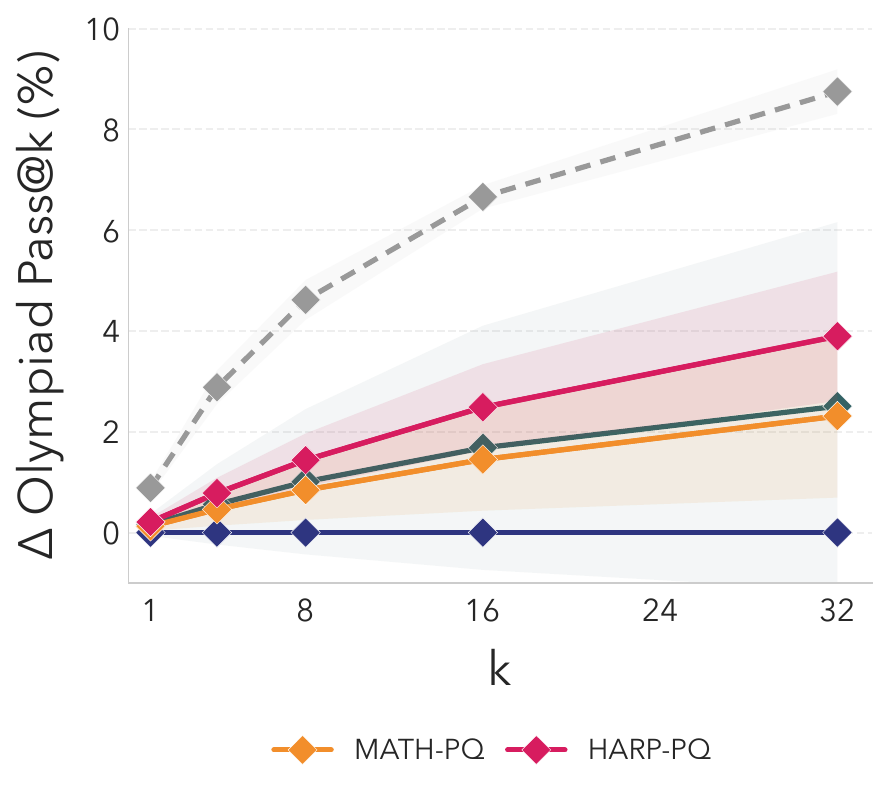}
        \caption{\textbf{Transfer performance to OlympiadBench fail@128 subset (improvement over \hardonly).} Questions optimized for MATH and HARP transfer to a held-out dataset. Absolute performance, including PS evaluation, is in Table \ref{tab:app-promotion-results-olympiad}. }
        \label{fig:olympiad-results}
    \end{minipage}
    \vspace{-1pt}
\end{figure*}

\subsection{Evaluation}
Once training completes, we test if the generated problems improve $\mathcal{D}_{test}$ performance. 
Based on observations of teacher reward plateaus in initial runs, we evaluate the teacher where training rewards stabilize: step 200 for MATH and step 170 for HARP. We assess two aspects of \algoname:

\textbf{Promoted Student (PS).} For training runs that reached multiple promotions, we evaluate the student model with the best validation performance (\textit{i.e.,} best $\mathcal{D}_{train}$ greedy accuracy) on the test set to measure direct performance gains from \algoname. In practice we observe a
maximum of four promotions; thus the PS model has
been trained on one of \{128, 192, 256\} synthetic questions.

\textbf{Promotion Questions (PQ).} We train a fresh base student with standard RLOO on a combination of PQ ($\mathcal{D}_{best}$) and the fail@128 train set. This isolates the value of
the synthetic questions, separate from the specific
training trajectory of the promoted student. We denote PQ from MATH and HARP training as PQ-MATH and PQ-HARP. In Appendix \ref{app:eval} we discuss how we mix synthetic and real data.

We compare to the following baselines:

\textbf{\hardonly.} We train directly on the $\mathcal{D}_{train}$ (real fail@128 train set) with a standard group size of 32. To disentangle the effects of the meta-RL loop from just using additional compute, we also train with group size 128 on MATH.

\textbf{Intrinsic Teacher (\intrinsic).}  To isolate the effects of grounding rewards, we compare to an intrinsic, data-free baseline. 
We train using the same procedure and hyperparameters as \algoname, but replace the grounded signal with a learnability objective \citep{zhao2025absolute, sukhbaatar2017asymmetric} that rewards questions of moderate difficulty.
We evaluate by training a fresh student on 128 problems sampled from this teacher (\intrinsic) alongside the fail@128 train set, following the PQ evaluation protocol. Learnability details in Appendix \ref{app:learnability}. 

\textbf{SeRL \citep{fang2025serl}.} SeRL serves as contemporary self-play baseline that self-evolves a curriculum from a seed set using self-rewards and diversity/learnability filters. We train SeRL with the MATH and HARP fail@128 train sets as the seed sets, until convergence. We use the hyperparameters specified by the authors for \texttt{Llama-3.2-3B-Instruct}. 

\textbf{Upper bound.} We train a fresh student the
full MATH train split (6750 problems) plus the fail@128 train set. This serves as an upper bound reference for performance with human-curated stepping stones.

\paragraph{Metrics.} We report the pass@k accuracy on the held-out fail@128 test set for $k \in \{1, 4, 8, 16, 32\}$, using 32 samples per problem. We report mean and standard deviation over 6-12 seeds, nested across teacher/student training (Appendix \ref{app:seeds}). For evaluations with fresh students, we do early stopping based on training reward convergence due to our small dataset size and differing convergence rates (Appendix \ref{app:eval}).

\begin{figure*}
\centering
    \includegraphics[width=1.0\textwidth]
    {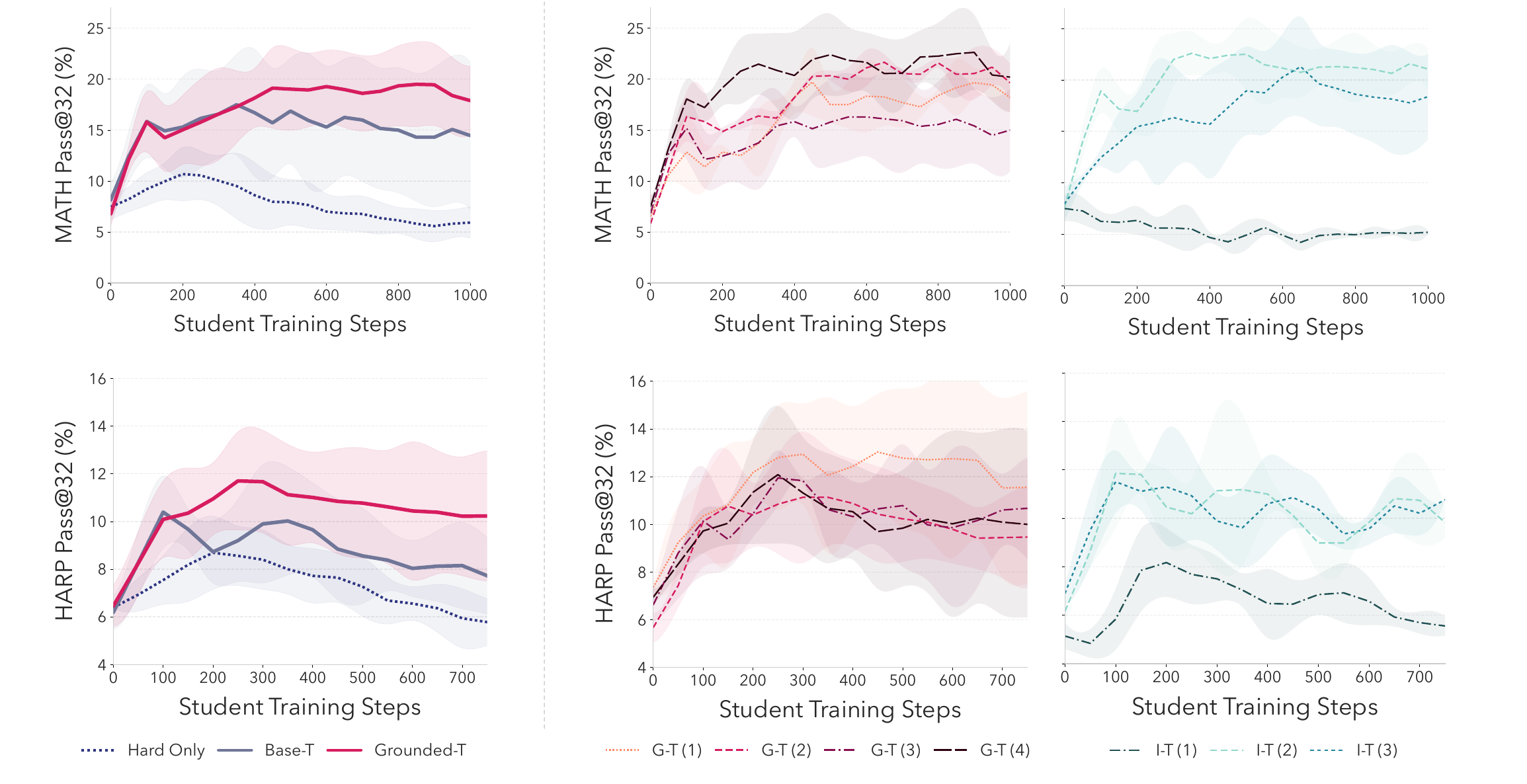}
     \caption{\textbf{Grounded rewards lead to more stable teacher policies.} We evaluate trained teacher policies by sampling questions and training fresh students. \textbf{(Left)} Test pass@32 comparison between students trained with questions sampled from \grounded and \base 
     (\hardonly also shown for reference). \grounded outperforms \base and exhibits more stable student trajectories. \textbf{(Right)} Pass@32 trajectories for fresh students trained with individual \grounded teacher seeds (red) and \intrinsic teacher seeds (green). Questions from \grounded yield consistent student trajectories, whereas \intrinsic exhibits higher variance across teachers, including a failure mode where I-T (1) causes student collapse. Shading shows $\pm 1$ SD. Curves for other pass@k and OlympiadBench are in Figures \ref{fig:teacher-ablation-MATH}-\ref{fig:teacher-ablation-olympiad}.}\label{fig:teacher-ablation}
     \vspace{-1pt}
\end{figure*}

\section{Results}

\subsection{Meta-RL Discovers Effective Questions.}

While curriculum learning is well-studied in RL, it is not obvious that synthetic questions can help a model move "beyond sharpening" its existing distributions.
We show that self-generated stepping stones provide a learnable gradient that unlocks improvement in stalled regimes.
This occurs without the teacher seeing the target problems; instead, meta-RL sharpens the teacher's policy, discovering useful curricula solely by optimizing for student progress.

\textbf{PQ kickstarts learning on hard subsets.} Both PS and PQ substantially outperform \hardonly and \intrinsic baselines, with larger gains at higher $k$. Figure \ref{fig:promotion-results} shows \textit{improvement over \hardonly}. \hardonly test trajectories are in Figures \ref{fig:teacher-ablation}; all absolute numbers and trajectories are in Appendix \ref{app:full-training-curves}-\ref{app:pq-full}. Inference with the base model achieves non-zero pass@$k$ due to stochastic sampling with different seeds than were used for the initial fail$@128$ filtering; nonetheless, \hardonly training plateaus. 

Inference with PS achieves +8.5\% pass@32 on fail@128-MATH and +3.6\% pass@32 on fail@128-HARP over \hardonly. PQ achieves higher mean performance (+9.3\% pass@32 on MATH, +4.2\% on HARP), indicating that \emph{the synthetic questions, rather than a fortunate student training trajectory, drive the performance gains.} Both \intrinsic and SeRL (Tables \ref{tab:app-promotion-results-math}-\ref{tab:app-promotion-results-harp}) perform worse across datasets, validating that {\em grounded rewards are needed to discover the right questions}. 

Synthetic questions also shift the student policy to make previously hard problems learnable; student learning curves on MATH show continued improvement after transitioning to real fail@128 training (Figure \ref{fig:accepted-qs-full}). These effects significantly outstrip what can be achieved from repeated sampling alone on fail@128 data. \hardonly with a group size of 128 (4$\times$ extra compute) achieves only +2.8\% pass@32 (Table \ref{tab:app-promotion-results-math}); furthermore, extending \hardonly training from 1500 to 6500 steps does not improve performance (Figure \ref{fig:harp-hardonly-extended}). 

\textbf{OOD generalization.} Figure \ref{fig:olympiad-results}
shows that synthetic questions from PQ-MATH, PQ-HARP, and \intrinsic \textit{transfer} to OlympiadBench, an OOD dataset. Cross-dataset transfer, despite no OOD optimization, suggests that synthetic curricula can capture generalizable reasoning pathways.

\textbf{Oracle comparison to real curated data.} 
Our regime assumes that we only have access to hard problems, to study the case where additional expert-curated data is not available or not known. As a strong upper-bound, we compare to the ``oracle" case where curated extra data is available. We train students on fail@128 + the full official MATH training set (6750 problems) as a representative pool of abundant, easier questions. We also compare to training with 128 random MATH/HARP questions in Appendix \ref{app:pq-full}, which performs similarly to training with the full dataset. Synthetic PQ-MATH questions recover 75\% of the performance gains from full-MATH training, and PQ-HARP recover 50\%. Notably, HARP-PQ ($128/192$ questions) outperforms 128 real HARP questions, and matches 128 real MATH questions. 

Direct inference on fail@128 test problems with the final \textit{trained teacher policy} model does not improve over base model performance (Appendix \ref{app:pq-full}), indicating that generator and solver abilities are largely independent.
\begin{takeaway}
\textbf{Takeaway:} 
A model’s \textit{pedagogical} ability can be decoupled from its \textit{task-solving} ability. Grounded meta-RL expands the ``learnability frontier" by surfacing synthetic questions that enable improving over reasoning plateaus.
\end{takeaway}

\begin{figure*}[h]
    \centering
    \includegraphics[width=1.0\linewidth]{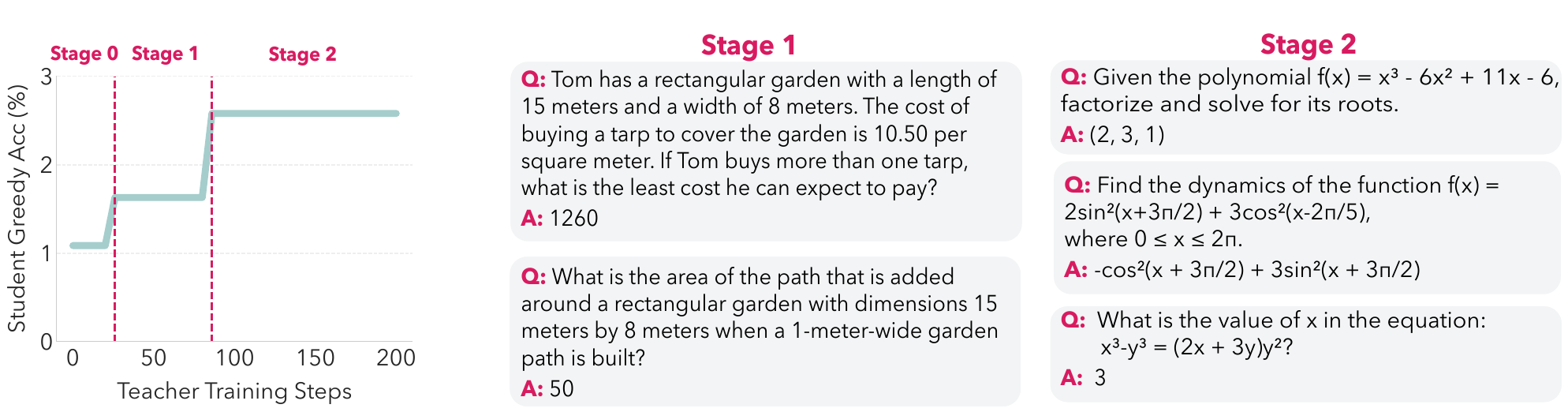}
    \caption{\textbf{Qualitative evolution of generated questions.} (Left) Baseline student performance during a \algoname run on HARP. The y-axis shows greedy accuracy on the \textit{fail@128 train set} over promotion stages. (Right) Sampled teacher questions at different promotion points. Content and style shift from word problems and basic formulas (stage 1) to concise, equation-heavy problems (stage 2). 
    Many effective ``stepping stones" include incorrect solutions, suggesting that structural and conceptual content provide sufficient learning signal.}
    \label{fig:qual}
    \vspace{-1pt}
\end{figure*}

\subsection{Grounded Rewards Yield Stable, Diverse Teachers.}
\label{sec:teacher-ablation}

While the main utility of \algoname is in surfacing a set of useful teacher-generated questions (PQ), we now shift focus to the trained teacher policies themselves.
We perform a controlled study of teacher objectives to probe the effects of meta-RL, and show that grounded rewards (as in \algoname), versus intrinsic ones, yield stronger teacher policies. We evaluate teachers trained with grounded rewards (\grounded), intrinsic rewards (\intrinsic) and the base model (\base) by sampling question-answer pairs from these policies and training fresh students. In Appendix \ref{app:teacher-ablation} we also ablate \algoname without the student-promotion mechanism, to validate its necessity. 

We evaluate four \grounded seeds per dataset to cover a range of final promotion stages and three \intrinsic teacher seeds. We sample 128 questions from each teacher and train 2-3 fresh students on the synthetic questions and real fail@128 train set ($\geq 9$ student runs per metric). 

\textbf{The teacher policy generates useful questions.} Student test performance curves in Figure \ref{fig:teacher-ablation} reveal that questions sampled from \grounded improve over \hardonly. Results are competitive with PQ on MATH and HARP, validating that the useful pedagogical signal is not just captured in the set of evolved questions, but is also learned by the teacher policy. Further ablations show that sampling larger datasets from \grounded reduces the variance of student outcomes (Appendix \ref{app:sampled-dataset-size}) and that the student-promotion mechanism improves the teacher policy (Appendix \ref{app:teacher-ablation}).

\textbf{Meta-RL sharpens the question distribution.} In Figure \ref{fig:teacher-ablation} (left) we overlay student training curves for \grounded questions and \base questions. \grounded students consistently track the upper envelope of \base performance for MATH/HARP, with lower variance on MATH. The existence of successful runs from \base reveals the ability to generate useful stepping stone questions is latent in the model; meta-RL improves \grounded by \textit{sharpening the teacher} to output questions that more reliably provide useful gradient signal. 
This is yet another example of the sharpening mechanism of RL \citep{yue2025does,zhao2025echo,tsilivis2025how}, but here leveraged for curricula. On OlympiadBench, where the target distribution differs substantially from the teacher’s training signal, \grounded and \base learning curves overlap more (though \grounded-HARP achieves best performance), suggesting that meta-RL primarily sharpens in-domain pedagogical signals. 

\textbf{Fragility of intrinsic proxies.} Figure \ref{fig:teacher-ablation} (right) compares aggregate student training curves for individual \grounded and \intrinsic teacher seeds. Students trained with questions from different \grounded seeds exhibit highly similar trajectories, indicating that grounded rewards lead to stable teacher policies. \intrinsic teachers produce, on average, worse and more volatile outcomes, with a clear separation in performance between students trained with different \intrinsic seeds across MATH, HARP, and OlympiadBench. 
While some \intrinsic teachers produce highly effective curricula, the objective is subject to a high-variance failure mode: one out of three teacher seeds exhibits collapse across all datasets, yielding little or no progress on the target problems. This reinforces observations from the literature  that RL with self-rewards is prone to reward hacking, or the decoupling of the intrinsic reward from actual task mastery \citep{shafayat2025largereasoningmodelsselftrain,chae2025understandingselfplay}.

\textbf{Grounded training sustains diversity.} 
To probe how meta-RL shapes the teacher’s generative distribution, in Table \ref{tab:final_diversity_results} we measure the semantic diversity of datasets from different teachers with the Vendi Score ($VS$) \citep{friedman2022vendi} using \texttt{Qwen3-8B embeddings} \citep{zhang2025qwen3}. \grounded (MATH) and \grounded (HARP) match the diversity of \base ($VS=34.91$), with PQ showing only a small decline from the base model ($VS = 31.75$). In contrast, \intrinsic collapses into a narrow conceptual space ($VS = 10.82$), providing evidence of reward-hacking and an explanation for the observed instability%
. This suggests that grounded rewards successfully avoid the diversity collapse often seen in RL-loops \citep{song2025outcomebasedexplorationllmreasoning}, while intrinsic rewards fall prey to it. Indeed, we also observe a decline in the diversity of teacher completions during meta-RL with learnability rewards (Appendix \ref{app:teacher-training-curves}).

\begin{table}[h] %
  \centering
  \caption{\textbf{Semantic diversity of synthetic datasets}. Diversity is measured with the Vendi Score ($VS$), representing the effective number of unique semantic concepts. All metrics are standardized to 128 questions via bootstrap subsampling ($k=100$ iterations). Our grounded reward (\grounded) preserves model diversity better than intrinsic rewards (\intrinsic).} \label{tab:final_diversity_results}
  \begin{tabular}{lcc}
    \toprule
    \textbf{Method} & \textbf{Vendi Score} & \textbf{Std. Dev} \\
     & ($VS$) & ($\sigma$) \\
    \midrule
    \base           & \textbf{34.91}          & \textbf{1.74} \\
    \grounded (HARP) & 34.66 & 1.74 \\
    \grounded (MATH) & 31.99 & 1.54 \\
    PQ              & 28.33          & 1.55 \\
    \intrinsic      & 10.82          & 1.01 \\
    \bottomrule
  \end{tabular}
\end{table}

\begin{takeaway}
\textbf{Takeaway:} 
Effective questions are latent in the base model, but hard to find. Grounding rewards in student progress "sharpens" the teacher's noisy distribution of questions into a stable, diversity-preserving policy, whereas intrinsic rewards are prone to instability and 
diversity collapse.
\end{takeaway}

\subsection{Question Structure over Answer Correctness. }
 While conventional wisdom suggests that question-answer correctness is most important, our results suggest that the \textit{conceptual content and structure of questions} is more important for models on learning plateaus. 

Figure \ref{fig:qual} shows qualitative examples of PQ questions at different stages of a sample \algoname training trajectory, exhibiting shifts in style and conceptual focus as the baseline student improves. We annotate synthetic questions with \texttt{Claude-4.5-Sonnet} as an oracle judge, and observe that only 32.8\% of PQ problems  contain a fully correct solution, while 63\% are considered mathematically well-posed (Appendix \ref{app:taxonomy}). This suggests that for models stalled on a performance plateau, structural and contextual cues of a question are more important for kickstarting learning than a correct answer. Indeed, \intrinsic questions have \textit{higher} correctness (55\%) but perform worse, likely because of lack of diversity (Section \ref{sec:teacher-ablation}). A more detailed taxonomy of error types is in Appendix \ref{app:taxonomy}. 
Meta-RL decreases question ambiguity errors relative to \base, validating the importance of question coherence over answer correctness. 

To better isolate these effects, we compare training on the subset of HARP-PQ questions with correct answers, to training on the well-posed subset (both correct and incorrect answers) in Appendix \ref{app:correctness-ablation}. Performance improves with the addition of well-posed questions with incorrect answers. Our experiments with \base (which, like \grounded and \intrinsic, is filtered for correctly formatted questions) show that question format alone is not behind these results.

\begin{takeaway}
    \textbf{Takeaway: } For models at learning plateaus, problems that have conceptually diverse and coherent \textit{questions} can provide useful gradient signal even without having precisely correct \textit{answers}.
\end{takeaway}

\section{Discussion and Conclusions}

\textbf{Breaking the sparse-reward plateau in RL fine-tuning.}
Our work establishes a way to kickstart RL fine-tuning when the initial success rate is too low to collect RLVR signal. Generating and training on question-answer pairs (even if not correct), with
the right meta-RL self-play loop, can be
enough to provide nonzero signal on the original hard problems.
Our setup shows that \textit{generating stepping-stone questions to solve a problem does not require the preexisting ability to solve that problem}, and that \textit{meta-RL sharpens this latent ability in the pretraining distribution}. 
A central contribution is that we show how to make this grounded bilevel meta-RL loop work in practice. The gap in performance shows the importance of this point.

\textbf{Grounded rewards mitigate self-play collapse.} The intuition of a gap between the abilities to generate and solve problems lies at the core of self-play. However, we show that it is crucial to go beyond pure curiosity by grounding in actual performance. Prior LLM self-play approaches use purely intrinsic rewards, such as learnability and self-consistency, to train the teacher. We find, however, that these objectives are prone to reward hacking, diversity collapse, and instability across seeds because they are decoupled from task performance. Grounding the teacher reward in measured student progress on real problems prevents teacher degeneration and preserves diversity.

\textbf{Expanding the learning frontier.} Our results tie to the broader debate on whether RL fine-tuning truly expands a model’s learning frontier, or merely sharpens latent abilities~\citep{yue2025does,zhao2025echo,tsilivis2025how}. Our work indicates that meta-RL can expand the envelope of learnability beyond what direct RLVR can achieve. As a “North Star” thought experiment, consider a future model trained on the entire mathematical literature: a proof of a Millennium Problem such as the Riemann Hypothesis may already be latent in pretraining, yet successful learning would hinge on recovering the right sequence of intermediate lemmas and theorems that make the proof \emph{learnable} to a student reasoner. In this view, just as RL is believed to amplify useful subsets of pretraining data, meta-RL could retrieve the stepping-stone question–answer pairs embedded in the teacher’s vast training corpus. We believe our results provide concrete evidence that a moderate amount of grounded meta-RL can elicit such capabilities that remain inaccessible through repeated sampling alone.

\ignore{More importantly, our setup shows that \textit{generating stepping-stones
questions to solve a problem does not require the preexisting ability to
solve that problem}. This intuition lies at the core of the self-play
idea, although we show it is important to  go beyond pure curiosity by
grounding in actual performance.

More generally, this is relevant for the discussion on whether or not RL
fine-tuning truly expands a model's learning frontier, or just sharpens
latent abilities
\citep{yue2025does,zhao2025echo,tsilivis2025how,tsilivis2025how}.
 As a ``North Star" thought experiment, consider a future model trained on the
entire mathematical literature: a proof of a Millennium Problem such as
the Riemann Hypothesis might already be latent in pretraining, yet
learning would hinge on recovering the right sequence of intermediate
lemmas and theorems that make the proof \emph{learnable} to a student
reasoner. And just like RL is believed to sharpen or amplify useful
subsets of the pretraining data, RL could also retrieve these stepping
stone question-answer pairs from the vast pretraining data of the teacher
LLM.
We believe our work shows that a moderate amount of
meta-RL training can bring out new abilities that could not be extracted
just by repeated sampling.}

\textbf{Limitations.} Our primary limitation is the computational cost of running bilevel RL loops (Appendix \ref{app:compute}). While inner loop training is relatively cheap (10-20 steps) it necessitates training parallel students for stability. Our ablations in Table \ref{tab:app-promotion-results-math} and Figure \ref{fig:harp-hardonly-extended} show that reallocating extra compute to direct training on hard problems does not recover the gains from meta-RL. Our work is a proof of concept for grounded rewards in this setting; more efficient reward proxies and scaling beyond 3-8B models are rich avenues for further work.

\ignore{\section*{Acknowledgements}
We thank Cansu Sancaktar and Phillip Isola for helpful discussions. 
JK thanks the Simons Foundation for support through the Collaborative Grant “The Physics of Learning
and Neural Computation”. 
This work was supported by an NSF GRFP fellowship to SS.}

\section*{Acknowledgements}
We thank Cansu Sancaktar, Reyhane Askari Hemmat, and Phillip Isola for helpful discussions. JK
thanks the Simons Foundation for support through the Collaborative Grant “The Physics of Learning
and Neural Computation”. This work was supported by an NSF GRFP fellowship to SS. This work
was also supported under project ID 43 as part of the Swiss AI Initiative, through a grant from the
ETH Domain and computational resources provided by the Swiss National Supercomputing Centre
(CSCS) under the Alps infrastructure

\newpage
\bibliography{reference}

@inproceedings{dennis2020paired,
	author = {Dennis, Michael and Jaques, Natasha and Vinitsky, Eugene and Bayen, Alexandre and Russell, Stuart and Critch, Andrew and Levine, Sergey},
	booktitle = {Advances in Neural Information Processing Systems},
	editor = {H. Larochelle and M. Ranzato and R. Hadsell and M.F. Balcan and H. Lin},
	pages = {13049--13061},
	publisher = {Curran Associates, Inc.},
	title = {Emergent Complexity and Zero-shot Transfer via Unsupervised Environment Design},
	url = {https://proceedings.neurips.cc/paper_files/paper/2020/file/985e9a46e10005356bbaf194249f6856-Paper.pdf},
	volume = {33},
	year = {2020}}

@article{song2025outcomebasedexplorationllmreasoning,
  title={Outcome-based Exploration for LLM Reasoning},
  author={Yuda Song and Julia Kempe and R{\'e}mi Munos},
  journal={ArXiv},
  year={2025},
  volume={abs/2509.06941},
  url={https://api.semanticscholar.org/CorpusID:281203824}
}

@inproceedings{
racaniere2020settersolver,
title={Automated curriculum generation through setter-solver interactions},
author={Sebastien Racaniere and Andrew Lampinen and Adam Santoro and David Reichert and Vlad Firoiu and Timothy Lillicrap},
booktitle={International Conference on Learning Representations},
year={2020},
url={https://openreview.net/forum?id=H1e0Wp4KvH}
}

@inproceedings{ahmadian-etal-2024-back,
    title = "Back to Basics: Revisiting {REINFORCE}-Style Optimization for Learning from Human Feedback in {LLM}s",
    author = {Ahmadian, Arash  and
      Cremer, Chris  and
      Gall{\'e}, Matthias  and
      Fadaee, Marzieh  and
      Kreutzer, Julia  and
      Pietquin, Olivier  and
      {\"U}st{\"u}n, Ahmet  and
      Hooker, Sara},
    editor = "Ku, Lun-Wei  and
      Martins, Andre  and
      Srikumar, Vivek",
    booktitle = "Proceedings of the 62nd Annual Meeting of the Association for Computational Linguistics (Volume 1: Long Papers)",
    month = aug,
    year = "2024",
    address = "Bangkok, Thailand",
    publisher = "Association for Computational Linguistics",
    url = "https://aclanthology.org/2024.acl-long.662/",
    doi = "10.18653/v1/2024.acl-long.662",
    pages = "12248--12267"
}

@article{
friedman2022vendi,
title={The Vendi Score: A Diversity Evaluation Metric for Machine Learning},
author={Dan Friedman and Adji Bousso Dieng},
journal={Transactions on Machine Learning Research},
issn={2835-8856},
year={2023},
url={https://openreview.net/forum?id=g97OHbQyk1},
note={}
}

@article{zhang2025qwen3,
  title={Qwen3 Embedding: Advancing Text Embedding and Reranking Through Foundation Models},
  author={Yanzhao Zhang and Mingxin Li and Dingkun Long and Xin Zhang and Huan Lin and Baosong Yang and Pengjun Xie and An Yang and Dayiheng Liu and Junyang Lin and Fei Huang and Jingren Zhou},
  journal={ArXiv},
  year={2025},
  volume={abs/2506.05176},
  url={https://api.semanticscholar.org/CorpusID:279243736}
}

@inproceedings{jiang2021ued,
	author = {Jiang, Minqi and Dennis, Michael and Parker-Holder, Jack and Foerster, Jakob and Grefenstette, Edward and Rockt\"{a}schel, Tim},
	booktitle = {Advances in Neural Information Processing Systems},
	editor = {M. Ranzato and A. Beygelzimer and Y. Dauphin and P.S. Liang and J. Wortman Vaughan},
	pages = {1884--1897},
	publisher = {Curran Associates, Inc.},
	title = {Replay-Guided Adversarial Environment Design},
	url = {https://proceedings.neurips.cc/paper_files/paper/2021/file/0e915db6326b6fb6a3c56546980a8c93-Paper.pdf},
	volume = {34},
	year = {2021}}

@inproceedings{
parashar2025curriculumreinforcementlearningeasy,
title={Curriculum Reinforcement Learning from Easy to Hard Tasks Improves {LLM} Reasoning},
author={Shubham Parashar and Shurui Gui and Xiner Li and Hongyi Ling and Sushil Vemuri and Blake Olson and Eric Li and Yu Zhang and James Caverlee and Dileep Kalathil and Shuiwang Ji},
booktitle={The Fourteenth International Conference on Learning Representations},
year={2026},
url={https://openreview.net/forum?id=KJvHnl3kUv}
}

@inproceedings{
jiang2025ADO,
title={Adaptive Data Optimization: Dynamic Sample Selection with Scaling Laws},
author={Yiding Jiang and Allan Zhou and Zhili Feng and Sadhika Malladi and J Zico Kolter},
booktitle={The Thirteenth International Conference on Learning Representations},
year={2025},
url={https://openreview.net/forum?id=aqok1UX7Z1}
}

@inproceedings{
wang2025stablellmselfplay,
title={Triplets Better Than Pairs: Towards Stable and Effective Self-Play Fine-Tuning for {LLM}s},
author={Yibo Wang and Hai-Long Sun and Guangda Huzhang and Qing-Guo Chen and Zhao Xu and Weihua Luo and Kaifu Zhang and Lijun Zhang},
booktitle={The Thirty-ninth Annual Conference on Neural Information Processing Systems},
year={2025},
url={https://openreview.net/forum?id=Hk4cCTukeI}
}

@article{kuba2025languageselfplay,
  title={Language Self-Play For Data-Free Training},
  author={Jakub Grudzien Kuba and Mengting Gu and Qi Ma and Yuandong Tian and Vijai Mohan},
  journal={ArXiv},
  year={2025},
  volume={abs/2509.07414},
  url={https://api.semanticscholar.org/CorpusID:281217919}
}

@inproceedings{wen2025lightr1,
    title = "Light-R1: Curriculum {SFT}, {DPO} and {RL} for Long {COT} from Scratch and Beyond",
    author = "Wen, Liang  and
      Cai, Yunke  and
      Xiao, Fenrui  and
      He, Xin  and
      An, Qi  and
      Duan, Zhenyu  and
      Du, Yimin  and
      Liu, Junchen  and
      Tang, Lifu  and
      Lv, Xiaowei  and
      Zou, Haosheng  and
      Deng, Yongchao  and
      Jia, Shousheng  and
      Zhang, Xiangzheng",
    editor = "Rehm, Georg  and
      Li, Yunyao",
    booktitle = "Proceedings of the 63rd Annual Meeting of the Association for Computational Linguistics (Volume 6: Industry Track)",
    month = jul,
    year = "2025",
    address = "Vienna, Austria",
    publisher = "Association for Computational Linguistics",
    url = "https://aclanthology.org/2025.acl-industry.24/",
    doi = "10.18653/v1/2025.acl-industry.24",
    pages = "318--327",
    ISBN = "979-8-89176-288-6",
}

@article{silver2018alphazero,
  title       = {A General Reinforcement Learning Algorithm that Masters Chess, Shogi, and Go through Self-Play},
  author      = {Silver, David and Hubert, Thomas and Schrittwieser, Julian and Antonoglou, Ioannis and Lai, Matthew H. and Guez, Arthur and Lanctot, Marc and Sifre, Laurent and Kumaran, Dharshan and Graepel, Thore and Lillicrap, Timothy and Simonyan, Karen and Hassabis, Demis},
  journal     = {\textit{Science}},
  volume      = {362},
  number      = {6419},
  pages       = {1140--1144},
  year        = {2018},
  doi         = {10.1126/science.aar6404},
  url         = {https://www.science.org/doi/10.1126/science.aar6404}
}

@inproceedings{kordi2025revisiting,
    title = "Revisiting Generalization Across Difficulty Levels: It{'}s Not So Easy",
    author = "Kordi, Yeganeh  and
      Nayak, Nihal V.  and
      Zuo, Max  and
      Nguyen, Ilana  and
      Bach, Stephen",
    editor = "Demberg, Vera  and
      Inui, Kentaro  and
      Marquez, Llu{\'i}s",
    booktitle = "Proceedings of the 19th Conference of the {E}uropean Chapter of the {A}ssociation for {C}omputational {L}inguistics (Volume 1: Long Papers)",
    month = mar,
    year = "2026",
    address = "Rabat, Morocco",
    publisher = "Association for Computational Linguistics",
    url = "https://aclanthology.org/2026.eacl-long.330/",
    doi = "10.18653/v1/2026.eacl-long.330",
    pages = "7014--7042",
    ISBN = "979-8-89176-380-7"
}

@inproceedings{he2024olympiadbenchchallengingbenchmarkpromoting,
    title = "{O}lympiad{B}ench: A Challenging Benchmark for Promoting {AGI} with Olympiad-Level Bilingual Multimodal Scientific Problems",
    author = "He, Chaoqun  and
      Luo, Renjie  and
      Bai, Yuzhuo  and
      Hu, Shengding  and
      Thai, Zhen  and
      Shen, Junhao  and
      Hu, Jinyi  and
      Han, Xu  and
      Huang, Yujie  and
      Zhang, Yuxiang  and
      Liu, Jie  and
      Qi, Lei  and
      Liu, Zhiyuan  and
      Sun, Maosong",
    editor = "Ku, Lun-Wei  and
      Martins, Andre  and
      Srikumar, Vivek",
    booktitle = "Proceedings of the 62nd Annual Meeting of the Association for Computational Linguistics (Volume 1: Long Papers)",
    month = aug,
    year = "2024",
    address = "Bangkok, Thailand",
    publisher = "Association for Computational Linguistics",
    url = "https://aclanthology.org/2024.acl-long.211/",
    doi = "10.18653/v1/2024.acl-long.211",
    pages = "3828--3850",
}

@article{AlphaProofNature2025,
	author = {Hubert, Thomas and Mehta, Rishi and Sartran, Laurent and Horv{\'a}th, Mikl{\'o}s Z. and {\v Z}u{\v z}i{\'c}, Goran and Wieser, Eric and Huang, Aja and Schrittwieser, Julian and Schroecker, Yannick and Masoom, Hussain and Bertolli, Ottavia and Zahavy, Tom and Mandhane, Amol and Yung, Jessica and Beloshapka, Iuliya and Ibarz, Borja and Veeriah, Vivek and Yu, Lei and Nash, Oliver and Lezeau, Paul and Mercuri, Salvatore and S{\"o}nne, Calle and Mehta, Bhavik and Davies, Alex and Zheng, Daniel and Pedregosa, Fabian and Li, Yin and von Glehn, Ingrid and Rowland, Mark and Albanie, Samuel and Velingker, Ameya and Schmitt, Simon and Lockhart, Edward and Hughes, Edward and Michalewski, Henryk and Sonnerat, Nicolas and Hassabis, Demis and Kohli, Pushmeet and Silver, David},
	date = {2025/11/12},
	doi = {10.1038/s41586-025-09833-y},
	id = {Hubert2025},
	isbn = {1476-4687},
	journal = {Nature},
	title = {Olympiad-level formal mathematical reasoning with reinforcement learning},
	url = {https://doi.org/10.1038/s41586-025-09833-y},
	year = {2025}}

@inproceedings{
chae2025understandingselfplay,
title={Towards Understanding Self-play for {LLM} Reasoning},
author={Justin Yang Chae and Md Tanvirul Alam and Nidhi Rastogi},
booktitle={The 5th Workshop on Mathematical Reasoning and AI at NeurIPS 2025},
year={2025},
url={https://openreview.net/forum?id=ENymEXXRVH}
}

@inproceedings{
sukhbaatar2017asymmetric,
title={Intrinsic Motivation and Automatic Curricula via Asymmetric Self-Play},
author={Sainbayar Sukhbaatar and Zeming Lin and Ilya Kostrikov and Gabriel Synnaeve and Arthur Szlam and Rob Fergus},
booktitle={International Conference on Learning Representations},
year={2018},
url={https://openreview.net/forum?id=SkT5Yg-RZ},
}

@misc{openai2021asymmetricselfplay,
      title={Asymmetric self-play for automatic goal discovery in robotic manipulation}, 
      author={OpenAI OpenAI and Matthias Plappert and Raul Sampedro and Tao Xu and Ilge Akkaya and Vineet Kosaraju and Peter Welinder and Ruben D'Sa and Arthur Petron and Henrique P. d. O. Pinto and Alex Paino and Hyeonwoo Noh and Lilian Weng and Qiming Yuan and Casey Chu and Wojciech Zaremba},
      year={2021},
      eprint={2101.04882},
      archivePrefix={arXiv},
      primaryClass={cs.LG},
      url={https://arxiv.org/abs/2101.04882}, 
}

@inproceedings{
sun2025rlgrokkingrecipe,
title={{RL} Grokking Recipe: How Does {RL} Unlock and Transfer New Algorithms in {LLM}s?},
author={Yiyou Sun and Yuhan Cao and Pohao Huang and Haoyue Bai and Hannaneh Hajishirzi and Nouha Dziri and Dawn Song},
booktitle={The Fourteenth International Conference on Learning Representations},
year={2026},
url={https://openreview.net/forum?id=CJJ8VxOWbG}
}

@inproceedings{
askari-hemmat2025improving,
title={Improving the Scaling Laws of Synthetic Data with Deliberate Practice},
author={Reyhane Askari-Hemmat and Mohammad Pezeshki and Elvis Dohmatob and Florian Bordes and Pietro Astolfi and Melissa Hall and Jakob Verbeek and Michal Drozdzal and Adriana Romero-Soriano},
booktitle={Forty-second International Conference on Machine Learning},
year={2025},
url={https://openreview.net/forum?id=0LZRtvK871}
}

@inproceedings{
zhao2025absolute,
title={Absolute Zero: Reinforced Self-play Reasoning with Zero Data},
author={Andrew Zhao and Yiran Wu and Yang Yue and Tong Wu and Quentin Xu and Yang Yue and Matthieu Lin and Shenzhi Wang and Qingyun Wu and Zilong Zheng and Gao Huang},
booktitle={The Thirty-ninth Annual Conference on Neural Information Processing Systems},
year={2025},
url={https://openreview.net/forum?id=neZSGqhxDa}
}

@misc{chen2025selfquestioning,
      title={Self-Questioning Language Models}, 
      author={Lili Chen and Mihir Prabhudesai and Katerina Fragkiadaki and Hao Liu and Deepak Pathak},
      year={2025},
      eprint={2508.03682},
      archivePrefix={arXiv},
      primaryClass={cs.LG},
      url={https://arxiv.org/abs/2508.03682}, 
}

@inproceedings{
chen2024spin,
title={Self-Play Fine-Tuning Converts Weak Language Models to Strong Language Models},
author={Zixiang Chen and Yihe Deng and Huizhuo Yuan and Kaixuan Ji and Quanquan Gu},
booktitle={Forty-first International Conference on Machine Learning},
year={2024},
url={https://openreview.net/forum?id=O4cHTxW9BS}
}

@inproceedings{ye2024eva,
  title={Scalable Reinforcement Post-Training Beyond Static Human Prompts: Evolving Alignment via Asymmetric Self-Play},
  author={Ziyu Ye and Rishabh Agarwal and Tianqi Liu and Rishabh Joshi and Sarmishta Velury and Quoc V. Le and Qijun Tan and Yuan Liu},
  year={2024},
  url={https://api.semanticscholar.org/CorpusID:273798270}
}

@inproceedings{
huang2025rzero,
title={R-Zero: Self-Evolving Reasoning {LLM} from Zero Data},
author={Chengsong Huang and Wenhao Yu and Xiaoyang Wang and Hongming Zhang and Zongxia Li and Ruosen Li and Jiaxin Huang and Haitao Mi and Dong Yu},
booktitle={The Fourteenth International Conference on Learning Representations},
year={2026},
url={https://openreview.net/forum?id=96apU6YzSO}
}

@inproceedings{
fang2025serl,
title={Se{RL}: Self-play Reinforcement Learning for Large Language Models with Limited Data},
author={Wenkai Fang and Shunyu Liu and Yang Zhou and Kongcheng Zhang and Tongya Zheng and Kaixuan Chen and Mingli Song and Dacheng Tao},
booktitle={The Thirty-ninth Annual Conference on Neural Information Processing Systems},
year={2025},
url={https://openreview.net/forum?id=ZF93vyH9He}
}

@article{chen2025sec,
  title={Self-Evolving Curriculum for LLM Reasoning},
  author={Xiaoyin Chen and Jiarui Lu and Minsu Kim and Dinghuai Zhang and Jian Tang and Alex Pich'e and Nicolas Gontier and Yoshua Bengio and Ehsan Kamalloo},
  journal={ArXiv},
  year={2025},
  volume={abs/2505.14970},
  url={https://api.semanticscholar.org/CorpusID:278782199}
}

@inproceedings{bae2025onlinedifficultyfilteringreasoning,
    title = "Online Difficulty Filtering for Reasoning Oriented Reinforcement Learning",
    author = "Bae, Sanghwan  and
      Hong, Jiwoo  and
      Lee, Min Young  and
      Kim, Hanbyul  and
      Nam, Jeongyeon  and
      Kwak, Donghyun",
    editor = "Demberg, Vera  and
      Inui, Kentaro  and
      Marquez, Llu{\'i}s",
    booktitle = "Proceedings of the 19th Conference of the {E}uropean Chapter of the {A}ssociation for {C}omputational {L}inguistics (Volume 1: Long Papers)",
    month = mar,
    year = "2026",
    address = "Rabat, Morocco",
    publisher = "Association for Computational Linguistics",
    url = "https://aclanthology.org/2026.eacl-long.30/",
    doi = "10.18653/v1/2026.eacl-long.30",
    pages = "700--719",
    ISBN = "979-8-89176-380-7"
}

@article{deepseek2025r1,
	author = {Guo, Daya and Yang, Dejian and Zhang, Haowei and Song, Junxiao and Wang, Peiyi and Zhu, Qihao and Xu, Runxin and Zhang, Ruoyu and Ma, Shirong and Bi, Xiao and others},
	date = {2025/09/01},
	doi = {10.1038/s41586-025-09422-z},
	id = {Guo2025},
	isbn = {1476-4687},
	journal = {Nature},
	number = {8081},
	pages = {633--638},
	title = {DeepSeek-R1 incentivizes reasoning in LLMs through reinforcement learning},
	url = {https://doi.org/10.1038/s41586-025-09422-z},
	volume = {645},
	year = {2025}}

@inproceedings{
tsilivis2025how,
title={How reinforcement learning after next-token prediction facilitates learning},
author={Nikolaos Tsilivis and Eran Malach and Karen Ullrich and Julia Kempe},
booktitle={The Fourteenth International Conference on Learning Representations},
year={2026},
url={https://openreview.net/forum?id=CTGpC7xWHM}
}

@article{
singh2024beyond,
title={Beyond Human Data: Scaling Self-Training for Problem-Solving with Language Models},
author={Avi Singh and John D Co-Reyes and Rishabh Agarwal and Ankesh Anand and Piyush Patil and Xavier Garcia and Peter J Liu and James Harrison and Jaehoon Lee and Kelvin Xu and Aaron T Parisi and others},
journal={Transactions on Machine Learning Research},
issn={2835-8856},
year={2024},
url={https://openreview.net/forum?id=lNAyUngGFK},
note={Expert Certification}
}

@inproceedings{bengio2009curriculum,
  title={Curriculum learning},
  author={Yoshua Bengio and J{\'e}r{\^o}me Louradour and Ronan Collobert and Jason Weston},
  booktitle={International Conference on Machine Learning},
  year={2009},
  url={https://api.semanticscholar.org/CorpusID:873046}
}

@InProceedings{Graves2017automatedcurriculum,
  title = 	 {Automated Curriculum Learning for Neural Networks},
  author =       {Alex Graves and Marc G. Bellemare and Jacob Menick and R{\'e}mi Munos and Koray Kavukcuoglu},
  booktitle = 	 {Proceedings of the 34th International Conference on Machine Learning},
  pages = 	 {1311--1320},
  year = 	 {2017},
  editor = 	 {Precup, Doina and Teh, Yee Whye},
  volume = 	 {70},
  series = 	 {Proceedings of Machine Learning Research},
  month = 	 {06--11 Aug},
  publisher =    {PMLR},
  url = 	 {https://proceedings.mlr.press/v70/graves17a.html},
}

@article{Navekar2020curriculum,
  author  = {Sanmit Narvekar and Bei Peng and Matteo Leonetti and Jivko Sinapov and Matthew E. Taylor and Peter Stone},
  title   = {Curriculum Learning for Reinforcement Learning Domains: A Framework and Survey},
  journal = {Journal of Machine Learning Research},
  year    = {2020},
  volume  = {21},
  number  = {181},
  pages   = {1--50},
  url     = {http://jmlr.org/papers/v21/20-212.html}
}

@article{kimiteam2025kimi,
  title={Kimi k1.5: Scaling Reinforcement Learning with LLMs},
  author={Kimi Team and Angang Du and Bofei Gao and Bowei Xing and Changjiu Jiang and Cheng Chen and Cheng Li and Chenjun Xiao and Chenzhuang Du and Chonghua Liao and others},
  journal={ArXiv},
  year={2025},
  volume={abs/2501.12599},
  url={https://api.semanticscholar.org/CorpusID:275789974}
}

@InProceedings{pmlr-v97-colas19a,
  title = 	 {{CURIOUS}: Intrinsically Motivated Modular Multi-Goal Reinforcement Learning},
  author =       {Colas, C{\'e}dric and Fournier, Pierre and Chetouani, Mohamed and Sigaud, Olivier and Oudeyer, Pierre-Yves},
  booktitle = 	 {Proceedings of the 36th International Conference on Machine Learning},
  pages = 	 {1331--1340},
  year = 	 {2019},
  editor = 	 {Chaudhuri, Kamalika and Salakhutdinov, Ruslan},
  volume = 	 {97},
  series = 	 {Proceedings of Machine Learning Research},
  month = 	 {09--15 Jun},
  publisher =    {PMLR},
  url = 	 {https://proceedings.mlr.press/v97/colas19a.html}
}

@inproceedings{
zhao2025learningreasonexternalrewards,
title={Learning to Reason without External Rewards},
author={Xuandong Zhao and Zhewei Kang and Aosong Feng and Sergey Levine and Dawn Song},
booktitle={The Fourteenth International Conference on Learning Representations},
year={2026},
url={https://openreview.net/forum?id=OU9nFEYR2M}
}

@article{prabhudesai2025maximizingconfidenceimprovesreasoning,
  title={Maximizing Confidence Alone Improves Reasoning},
  author={Mihir Prabhudesai and Lili Chen and Alex Ippoliti and Katerina Fragkiadaki and Hao Liu and Deepak Pathak},
  journal={ArXiv},
  year={2025},
  volume={abs/2505.22660},
  url={https://api.semanticscholar.org/CorpusID:278959314}
}

@article{shafayat2025largereasoningmodelsselftrain,
  title={Can Large Reasoning Models Self-Train?},
  author={Sheikh Shafayat and Fahim Tajwar and Ruslan Salakhutdinov and Jeff G. Schneider and Andrea Zanette},
  journal={ArXiv},
  year={2025},
  volume={abs/2505.21444},
  url={https://api.semanticscholar.org/CorpusID:278911518}
}

@inproceedings{
zuo2025ttrl,
title={{TTRL}: Test-Time Reinforcement Learning},
author={Yuxin Zuo and Kaiyan Zhang and Li Sheng and Shang Qu and Ganqu Cui and Xuekai Zhu and Haozhan Li and Yuchen Zhang and Xinwei Long and Ermo Hua and Biqing Qi and Youbang Sun and Zhiyuan Ma and Lifan Yuan and Ning Ding and Bowen Zhou},
booktitle={The Thirty-ninth Annual Conference on Neural Information Processing Systems},
year={2025},
url={https://openreview.net/forum?id=VuVhgEiu20}
}

@inproceedings{
feng2024embarrassingly,
title={Embarrassingly Simple Dataset Distillation},
author={Yunzhen Feng and Shanmukha Ramakrishna Vedantam and Julia Kempe},
booktitle={The Twelfth International Conference on Learning Representations},
year={2024},
url={https://openreview.net/forum?id=PLoWVP7Mjc}
}

@inproceedings{nguyen2021kipimprovedresults,
	author = {Nguyen, Timothy and Novak, Roman and Xiao, Lechao and Lee, Jaehoon},
	booktitle = {Advances in Neural Information Processing Systems},
	editor = {M. Ranzato and A. Beygelzimer and Y. Dauphin and P.S. Liang and J. Wortman Vaughan},
	pages = {5186--5198},
	publisher = {Curran Associates, Inc.},
	title = {Dataset Distillation with Infinitely Wide Convolutional Networks},
	url = {https://proceedings.neurips.cc/paper_files/paper/2021/file/299a23a2291e2126b91d54f3601ec162-Paper.pdf},
	volume = {34},
	year = {2021}}

@inproceedings{zhou2022dataset,
	author = {Zhou, Yongchao and Nezhadarya, Ehsan and Ba, Jimmy},
	booktitle = {Advances in Neural Information Processing Systems},
	editor = {S. Koyejo and S. Mohamed and A. Agarwal and D. Belgrave and K. Cho and A. Oh},
	pages = {9813--9827},
	publisher = {Curran Associates, Inc.},
	title = {Dataset Distillation using Neural Feature Regression},
	url = {https://proceedings.neurips.cc/paper_files/paper/2022/file/3fe2a777282299ecb4f9e7ebb531f0ab-Paper-Conference.pdf},
	volume = {35},
	year = {2022}}

@inproceedings{deng2022remember,
	author = {Deng, Zhiwei and Russakovsky, Olga},
	booktitle = {Advances in Neural Information Processing Systems},
	editor = {S. Koyejo and S. Mohamed and A. Agarwal and D. Belgrave and K. Cho and A. Oh},
	pages = {34391--34404},
	publisher = {Curran Associates, Inc.},
	title = {Remember the Past: Distilling Datasets into Addressable Memories for Neural Networks},
	url = {https://proceedings.neurips.cc/paper_files/paper/2022/file/de3d2bb604cfc43c81edd2a31b257f03-Paper-Conference.pdf},
	volume = {35},
	year = {2022}}

@inproceedings{maclaurin2015hyperopt,
  title     = {Gradient-based Hyperparameter Optimization through Reversible Learning},
  author    = {Maclaurin, Dougal and Duvenaud, David and Adams, Ryan P.},
  booktitle = {Proceedings of the 32nd International Conference on Machine Learning (ICML)},
  year      = {2015},
  pages     = {2113--2122},
  url       = {https://proceedings.mlr.press/v37/maclaurin15.html}
}

@misc{duan2016rl2fastreinforcementlearning,
      title={RL$^2$: Fast Reinforcement Learning via Slow Reinforcement Learning}, 
      author={Yan Duan and John Schulman and Xi Chen and Peter L. Bartlett and Ilya Sutskever and Pieter Abbeel},
      year={2016},
      eprint={1611.02779},
      archivePrefix={arXiv},
      primaryClass={cs.AI},
      url={https://arxiv.org/abs/1611.02779}, 
}

@InProceedings{Finn17maml,
  title = 	 {Model-Agnostic Meta-Learning for Fast Adaptation of Deep Networks},
  author =       {Chelsea Finn and Pieter Abbeel and Sergey Levine},
  booktitle = 	 {Proceedings of the 34th International Conference on Machine Learning},
  pages = 	 {1126--1135},
  year = 	 {2017},
  editor = 	 {Precup, Doina and Teh, Yee Whye},
  volume = 	 {70},
  series = 	 {Proceedings of Machine Learning Research},
  month = 	 {06--11 Aug},
  publisher =    {PMLR},
  url = 	 {https://proceedings.mlr.press/v70/finn17a.html}
}

@article{nichol2018firstordermetalearningalgorithms,
  title={On First-Order Meta-Learning Algorithms},
  author={Alex Nichol and Joshua Achiam and John Schulman},
  journal={ArXiv},
  year={2018},
  volume={abs/1803.02999},
  url={https://api.semanticscholar.org/CorpusID:4587331}
}

@inproceedings{
zhao2025echo,
title={Echo Chamber: {RL} Post-training Amplifies Behaviors Learned in Pretraining},
author={Rosie Zhao and Alexandru Meterez and Sham M. Kakade and Cengiz Pehlevan and Samy Jelassi and Eran Malach},
booktitle={Second Conference on Language Modeling},
year={2025},
url={https://openreview.net/forum?id=dp4KWuSDzj}
}

@inproceedings{
yu2025dapo,
title={{DAPO}: An Open-Source {LLM} Reinforcement Learning System at Scale},
author={Qiying Yu and Zheng Zhang and Ruofei Zhu and Yufeng Yuan and Xiaochen Zuo and YuYue and Weinan Dai and Tiantian Fan and Gaohong Liu and Juncai Liu and LingJun Liu and others},
booktitle={The Thirty-ninth Annual Conference on Neural Information Processing Systems},
year={2025},
url={https://openreview.net/forum?id=2a36EMSSTp}
}

@inproceedings{
hendrycks2021measuring,
title={Measuring Mathematical Problem Solving With the {MATH} Dataset},
author={Dan Hendrycks and Collin Burns and Saurav Kadavath and Akul Arora and Steven Basart and Eric Tang and Dawn Song and Jacob Steinhardt},
booktitle={Thirty-fifth Conference on Neural Information Processing Systems Datasets and Benchmarks Track (Round 2)},
year={2021},
url={https://openreview.net/forum?id=7Bywt2mQsCe}
}

@article{yue2024harp,
  title={HARP: A challenging human-annotated math reasoning benchmark},
  author={Albert S. Yue and Lovish Madaan and Ted Moskovitz and DJ Strouse and Aaditya K. Singh},
  journal={ArXiv},
  year={2024},
  volume={abs/2412.08819},
  url={https://api.semanticscholar.org/CorpusID:274655941}
}

@article{Mahsereci2017EarlySW,
  title={Early Stopping without a Validation Set},
  author={Maren Mahsereci and Lukas Balles and Christoph Lassner and Philipp Hennig},
  journal={ArXiv},
  year={2017},
  volume={abs/1703.09580},
  url={https://api.semanticscholar.org/CorpusID:14520242}
}

@inproceedings{
yue2025does,
title={Does Reinforcement Learning Really Incentivize Reasoning Capacity in {LLM}s Beyond the Base Model?},
author={Yang Yue and Zhiqi Chen and Rui Lu and Andrew Zhao and Zhaokai Wang and Yang Yue and Shiji Song and Gao Huang},
booktitle={The Thirty-ninth Annual Conference on Neural Information Processing Systems},
year={2025},
url={https://openreview.net/forum?id=4OsgYD7em5}
}

@InProceedings{pathak2017curiosity,
  title = 	 {Curiosity-driven Exploration by Self-supervised Prediction},
  author =       {Deepak Pathak and Pulkit Agrawal and Alexei A. Efros and Trevor Darrell},
  booktitle = 	 {Proceedings of the 34th International Conference on Machine Learning},
  pages = 	 {2778--2787},
  year = 	 {2017},
  editor = 	 {Precup, Doina and Teh, Yee Whye},
  volume = 	 {70},
  series = 	 {Proceedings of Machine Learning Research},
  month = 	 {06--11 Aug},
  publisher =    {PMLR},
  url = 	 {https://proceedings.mlr.press/v70/pathak17a.html}
}

@article{wang2018dataset,
  title={Dataset distillation},
  author={Wang, Tongzhou and Zhu, Jun-Yan and Torralba, Antonio and Efros, Alexei A},
  journal={arXiv preprint arXiv:1811.10959},
  year={2018},
  url={https://arxiv.org/abs/1811.10959}
}

@inproceedings{Jiang2020PrioritizedLR,
  title={Prioritized Level Replay},
  author={Minqi Jiang and Edward Grefenstette and Tim Rockt{\"a}schel},
  booktitle={International Conference on Machine Learning},
  year={2020},
  url={https://api.semanticscholar.org/CorpusID:222208809}
}

@inproceedings{
zweiger2025selfadapting,
title={Self-Adapting Language Models},
author={Adam Zweiger and Jyothish Pari and Han Guo and Yoon Kim and Pulkit Agrawal},
booktitle={The Thirty-ninth Annual Conference on Neural Information Processing Systems},
year={2025},
url={https://openreview.net/forum?id=JsNUE84Hxi}
}

@inproceedings{Schmidhuber1991APF,
author = {Schmidhuber, J\"{u}rgen},
title = {A possibility for implementing curiosity and boredom in model-building neural controllers},
year = {1991},
isbn = {0262631385},
publisher = {MIT Press},
address = {Cambridge, MA, USA},
booktitle = {Proceedings of the First International Conference on Simulation of Adaptive Behavior on From Animals to Animats},
pages = {222–227},
numpages = {6},
location = {Paris, France}
}

@inproceedings{NEURIPS2019_b6f97e6f,
	author = {Blaes, Sebastian and Vlastelica Pogan\v{c}i\'{c}, Marin and Zhu, Jiajie and Martius, Georg},
	booktitle = {Advances in Neural Information Processing Systems},
	editor = {H. Wallach and H. Larochelle and A. Beygelzimer and F. d\textquotesingle Alch\'{e}-Buc and E. Fox and R. Garnett},
	publisher = {Curran Associates, Inc.},
	title = {Control What You Can: Intrinsically Motivated Task-Planning Agent},
	url = {https://proceedings.neurips.cc/paper_files/paper/2019/file/b6f97e6f0fd175613910d613d574d0cb-Paper.pdf},
	volume = {32},
	year = {2019}}

@inproceedings{sancaktarRegularity,
	author = {Sancaktar, Cansu and Piater, Justus and Martius, Georg},
	booktitle = {Advances in Neural Information Processing Systems},
	editor = {A. Oh and T. Naumann and A. Globerson and K. Saenko and M. Hardt and S. Levine},
	pages = {62601--62629},
	publisher = {Curran Associates, Inc.},
	title = {Regularity as Intrinsic Reward for Free Play},
	url = {https://proceedings.neurips.cc/paper_files/paper/2023/file/c529dba08a146ea8d6cf715ae8930cbe-Paper-Conference.pdf},
	volume = {36},
	year = {2023}}

@article{colasAutotelic,
author = {Colas, C\'{e}dric and Karch, Tristan and Sigaud, Olivier and Oudeyer, Pierre-Yves},
title = {Autotelic Agents with Intrinsically Motivated Goal-Conditioned Reinforcement Learning: A Short Survey},
year = {2022},
issue_date = {Sep 2022},
publisher = {AI Access Foundation},
address = {El Segundo, CA, USA},
volume = {74},
issn = {1076-9757},
url = {https://doi.org/10.1613/jair.1.13554},
doi = {10.1613/jair.1.13554},
journal = {J. Artif. Int. Res.},
month = sep,
numpages = {41}
}

@misc{kydlicek2025mathverify,
  author = {Kydl{\'i}{\v{c}}ek, Hynek},
  title = {Math-Verify: Math Verification Library},
  year = {2025},
  publisher = {GitHub},
  journal = {GitHub repository},
  howpublished = {\url{https://github.com/huggingface/math-verify}}
}
\bibliographystyle{plainnat}
\newpage
\appendix

\etocsetnexttocdepth{subsection}
\etocsettocstyle{\section*{Appendix}}

\begingroup
\makeatletter
\newcommand{\ifappendixentry}[2]{%
  \begingroup
  \edef\@tempa{\etocthenumber}%
  \expandafter\@checkfirstchar\@tempa\@nil
  \if@isappendix
    \endgroup #1%
  \else
    \endgroup #2%
  \fi
}
\newif\if@isappendix
\def\@checkfirstchar#1#2\@nil{%
  \@isappendixfalse
  \ifnum`#1>`@ \ifnum`#1<`[ \@isappendixtrue \fi\fi
}
\makeatother

\etocsetstyle{section}
  {}
  {\leavevmode\leftskip=0pt\relax}
  {\ifappendixentry
    {\bfseries\makebox[2em][l]{\etocnumber}\etocname\nobreak\leaders\hbox{.}\hfill\nobreak\etocpage\par\vspace{0.5em}}
    {}}
  {}
\etocsetstyle{subsection}
  {}
  {\leavevmode\leftskip=2em\relax}
  {\ifappendixentry
    {\makebox[2.5em][l]{\etocnumber}\etocname\nobreak\leaders\hbox{.}\hfill\nobreak\etocpage\par\vspace{0.5em}}
    {}}
  {}

\etocsettocdepth{subsection}
\etocignoretoctocdepth
\tableofcontents
\endgroup

\section{Extended Related Work}\label{app:related}

\subsection{Curriculum Learning in RL}

Automated curriculum design has a long history predating modern LLMs,
beginning with classical curriculum learning
\citep{bengio2009curriculum,Graves2017automatedcurriculum}. These methods
assume access to a labeled training set and focus on \emph{reordering} or
\emph{selecting} existing data rather than generating new tasks. In the
context of RL, curriculum learning helps agents acquire complex behaviors
by first mastering simpler tasks \citep{Navekar2020curriculum,parashar2025curriculumreinforcementlearningeasy}, or those with high future learning potential \citep{Jiang2020PrioritizedLR}.
Contemporary LLM post-training inherits this paradigm: curriculum is
applied over curated prompts or problem categories, using proxy signals
such as gradient norms or advantage estimates to guide selection.
Examples include synthetic or self-training curricula like
Kimi~\citep{kimiteam2025kimi}, FastCuRL~\citep{dennis2020paired}, and
LightR1~\citep{wen2025lightr1}, as well as online difficulty-filtering
strategies such as Dapo~\citep{yu2025dapo}, Online Difficulty
Filtering~\citep{bae2025onlinedifficultyfilteringreasoning}, and
SEC~\citep{chen2025sec}, which discretize problems into difficulty
buckets and score categories by gradient-derived proxies. While these
approaches improve learning efficiency in-distribution or OOD, they
presuppose that difficulty can be meaningfully partitioned \emph{a
priori} and provide only indirect rewards for student progress. Adaptive Data Optimization (ADO) \citep{jiang2025ADO} leverages per-domain scaling laws to estimate the learning potential of various data sources online \cite{jiang2025ADO}.
By contrast, our goal is not to arrange data but to elicit
learning on a fixed, verifiable hard dataset where standard GRPO fails.

Another line of work explores unsupervised curriculum generation and environment design, moving beyond static datasets by generating a distribution of novel environments suitable for the agent's capabilities \citep{dennis2020paired,racaniere2020settersolver,Jiang2020PrioritizedLR,jiang2021ued}. Notably, these works find that unconstrained objectives such as minimax adversarial objectives lead to degenerate curricula and reward hacking, and instead optimize regret or estimates of learning potential. We diverge by grounding rewards in student progress on a fixed target set.

\subsection{Self-Play and Teacher-Student Setups}

Self-play offers a complementary lens on autonomous capability growth, classically exemplified by game-playing agents trained without external data, such as AlphaZero~\citep{silver2018alphazero}. Our approach is inspired by a  line of research demonstrating that {\em asymmetric} self-play can induce powerful automatic curricula. In early work, ~\citet{sukhbaatar2017asymmetric} introduced the canonical Alice--Bob framework in which one agent (Alice) proposes tasks while another (Bob) attempts to solve them, yielding a natural progression of ``just-hard-enough'' challenges that drive learning. This idea was later extended to complex embodied domains in robotics, where asymmetric self-play enabled automatic discovery of diverse manipulation goals without manual task specification~\citep{openai2021asymmetricselfplay}. 
Applying these ideas from robotics and control to large language models introduces fundamentally different challenges: LLMs operate over a discrete, symbolic problem space with no environment simulator to evaluate intermediate progress; a teacher must generate \emph{entire tasks}, often requiring multi-step reasoning. Moreover, rewards in language domains are extremely sparse and brittle---for mathematical problems, correctness is essentially binary and offers no gradient toward partial solutions. 
Modern LLM self-play methods thus differ in mechanism: SPIN \citep{chen2024spin}, Triplet self-play \citep{wang2025stablellmselfplay}, and ReST$^{\text{EM}}$ \citep{singh2024beyond} optimize for self-consistency or solution quality. These methods generate responses and still presuppose the existence
of well-formed input prompts or curated high-quality questions. Recent systems like AlphaProof \citep{AlphaProofNature2025} attempt to mitigate this sparsity at test-time by using an LLM to generate a "natural curriculum" of auxiliary theorem variations for additional training \citep{AlphaProofNature2025}. In the context of RLHF, eva~\citep{ye2024eva} casts RLHF as an asymmetric creator--solver game in which a creator evolves prompts to expose alignment weaknesses and a solver adapts to reward-model feedback.

A series of near-contemporary works leverages pre-trained LLMs themselves as an untapped resource for question generation.
Such "fully data-free" co-evolving systems—including Absolute Zero
\citep{zhao2025absolute}, R-Zero \citep{huang2025rzero}, Language
Self-Play (LSP) \citep{kuba2025languageselfplay}, SeRL
\citep{fang2025serl} and Self-Questioning Language Models (SQLM)
\citep{chen2025selfquestioning}—jointly evolve task creators and solvers
via intrinsic or proxy rewards such as majority vote, learnability,
reward-model preferences, or gradient magnitudes. Because these methods
optimize intrinsic or proxy objectives, they risk drifting to degenerate
or unlearnable tasks, are sensitive to reward hacking where models learn to maximize training
(pseudo-)reward, and lack guarantees
of progress (see an analysis of AbsoluteZero in
\citet{chae2025understandingselfplay}). This connects directly to a line of works investigating the broader question of whether self-training — the process where a model
learns from its own judgments — can be sustained within RL, and how far self-improvement can be driven by intrinsic or self-generated rewards.
Prolonged RL with self-rewards often results in sudden and complete performance collapse \citep{shafayat2025largereasoningmodelsselftrain,chae2025understandingselfplay}, when rewards vanish or when generator and solver objectives misalign, especially in discrete, symbolic domains with essentially binary correctness signals. 
Earlier
findings in unsupervised curriculum generation likewise note that proxy rewards (such as minimax objectives) can be fragile
\citep{dennis2020paired}. These
observations motivate our design: we learn a teacher \emph{policy} via
meta-RL that generates verifiable math questions directly optimized for
student learning progress, grounding the curriculum in a concrete failure
regime instead of internal proxy of difficulty.

\subsection{Intrinsic Rewards versus Bilevel Optimization} 

The use of intrinsic rewards is rooted in past studies across robotics, simulation, and task-agnostic settings for curricula generation
and exploration. Such rewards maximize objectives such as estimated learning progress, regularity, and prediction error   \cite{Schmidhuber1991APF,pathak2017curiosity,pmlr-v97-colas19a,NEURIPS2019_b6f97e6f,colasAutotelic,sancaktarRegularity}.  To our knowledge, essentially all recent ``fully data-free'' self-play approaches use
intrinsic or proxy rewards to train the teacher/proposer, without
anchoring to ``real'' student performance (with the exception of the self-adaptation work by \citet{zweiger2025selfadapting} which uses ReST$^{\text{EM}}$/SFT for outer/inner loop). 
Examples of intrinsic rewards include model confidence as proposed in Inuitor \citep{zhao2025learningreasonexternalrewards}, prediction entropy in DP \citep{askari-hemmat2025improving}, RENT \citep{prabhudesai2025maximizingconfidenceimprovesreasoning}, or the majority answer as in TTRL \citep{zuo2025ttrl} or \citet{shafayat2025largereasoningmodelsselftrain}, as well as in SQLM \citep{chen2025selfquestioning}. 

Of course, the use of proxy rewards is often not merely a design
preference but a pragmatic simplification, especially in teacher-student
self-play setups: it avoids facing an explicit inner-loop–outer-loop bilevel optimization problem - an appealing but  challenging objective where the output of one optimization (in this instance the optimization of the student trained with RLVF on the teacher's question-answer pairs) is fed into another optimization loop (the performance improvement of the student on the hard dataset). 
Such bilevel optimization objectives have strong historical precedence in
meta-learning, in popular methods such as MaML \citep{Finn17maml} and
Reptile \citep{nichol2018firstordermetalearningalgorithms}, which
explicitly train through an inner-loop–outer-loop structure to obtain
efficient few-shot learners,
following earlier research like RL2 \citep{duan2016rl2fastreinforcementlearning},
and works that meta-learn hyperparameters
of neural nets via full backpropagation through the training loop
\citep{maclaurin2015hyperopt}. 

A similar bilevel formulation, which
served as inspiration for our work, also appears in dataset distillation
\citep{wang2018dataset}, where an outer loop optimizes a generally small
dataset that allows an inner training loop to achieve good target
performance. Here, both proxy-based (e.g., NTK approximation
\citep{nguyen2021kipimprovedresults} or feature-matching
\citep{zhou2022dataset}) and end-to-end bilevel formulations have been
explored \citep{wang2018dataset,
deng2022remember,feng2024embarrassingly}. In general, such approaches become intractable, as the inner loop involves a multi-step computation
with a large number of steps, which requires backpropagation through time
(BPTT), or ``backpropagation through gradient descent'',
unrolling the inner loop and taking meta-gradients. Our approach,
however, avoids the need to unroll the inner loop thanks to the use of
RLOO in the outer loop,  using the reward (the performance improvement of
the student) to reinforce question-answer sets. This is the first
instance of a ``double meta-RL loop'' we are aware of in the context of self-play for LLMs.

\section{Method and Experiment Details}

\subsection{Prompts}\label{app:prompts}

\paragraph{Teacher Prompt.}
At every outer-loop step, the teacher is given the same prompt. The prompt guides the model towards producing valid math problems using sample subjects/domains and provides explicit instruction regarding the expected format. We avoid seeding the teacher with sample math questions to preserve the data-free setup; the model only sees the black-box reward signal of student performance. We also observe in initial experiments that, when given seed questions, the diversity of teacher generations decreases significantly, and the teacher typically collapses to copying the seed questions.

\begin{tcblisting}{
  arc=0mm,
  outer arc=0mm,
  colback=gray!5,
  colframe=gray!50,
  listing only, 
  listing options={style=tcblatex, breaklines=true, basicstyle=\ttfamily\small},
  title=Teacher Prompt
}
You are generating a new math problem for a math assistant.

Allowed topics: Algebra, Counting and Probability, Geometry, Intermediate Algebra, Number Theory, Prealgebra, or Precalculus.


Output rules (follow EXACTLY):
    - Provide the final formatted problem in this structure: <question>[full math question]<question><answer>\\boxed{[answer]}</answer>
    - Any explanations, steps, or reasoning about the problem goes OUTSIDE the <question> and <answer> tags.
    
    Constraints: 
        - The problem must be original, challenging, and require at least 2--3 steps of reasoning. 
        - Output exactly ONE problem. You MUST follow the specified format EXACTLY.
Begin now:
\end{tcblisting}

\paragraph{Student Prompt.}
The same prompt is used for fail@128 filtering, training the student in the inner-loop, and training the student in evaluation. 

\begin{tcblisting}{
  arc=0mm,
  outer arc=0mm,
  colback=gray!5,
  colframe=gray!50,
  listing only, 
  listing options={style=tcblatex, breaklines=true, basicstyle=\ttfamily\small},
  title=Student Prompt
}
A conversation between User and Assistant. The user asks a question, and the Assistant solves it. The assistant first shows the complete reasoning process step by step, then provides the final answer in \\boxed{}. The assistant must always follow the format: 'User: [question] Assistant: [detailed reasoning] The final answer is: \\boxed{[answer]}.
User: <QUESTION> Assistant: "
\end{tcblisting}

\subsection{Parsing Teacher Outputs}
\label{sec:filtering}

To parse the teacher
rollouts into question-answer pairs, we require teacher responses to
follow the prompt-specified format. We filter out generations that do not
follow this format, and resample until we have $g \cdot n$
correctly-formatted problems. We filter for the following:
\begin{itemize}
    \item Contains opening and closing question/answer tags.
    \item Contains the ``boxed" notation (denoting an answer).
    \item Contents of the boxed answer are parsable by a symbolic math verifier.
\end{itemize}

Theoretically, rejection sampling does not
affect the RLOO gradient update (Proposition~\ref{prop:RLOOfilter}); empirically, we find that this performs
better than using teacher-format rewards or sequential question/answer
sampling.

\begin{prop}[RLOO update with rejection sampling]
\label{prop:RLOOfilter}
Let $\pi_0(z)$ be a proposal distribution over some random variable $z$.
Let $S$ be a set of ``accepted'' values of $z$, and assume $\pi_0(S)>0$.
Let
\begin{equation}
\pi(z)= \pi_0(z)1_{z\in S}/\pi_0(S)
\end{equation}
be the distribution on $z$ obtained by rejection sampling, namely,
sampling $z$ from $\pi_0$ until $z\in S$.

Let $R(z)$ be some reward function on $z$. Then the RLOO update on $\pi$
can be computed from gradient of $\pi_0$ only. Namely, for any $g$-tuple
$z_1,\ldots,z_g$ sampled from $\pi$, one has
\begin{equation}
\sum_{i=1}^g A(z_i)\nabla \ln \pi(z_i)=\sum_{i=1}^g A(z_i)\nabla \ln
\pi_0(z_i)
\end{equation}
where
\begin{equation}
A(z_i)=R(z_i)-\frac{1}{g-1}\sum_{j\neq i} R(z_j)
\end{equation}
is the RLOO advantage function, and where the gradients are with respect
to the parameters of $\pi$.
\end{prop}

This is not true for simple Reinforce: it relies on the fact that RLOO
advantages $A(z_i)$ sum to $0$ over $i$.

\begin{proof}
For any $z$ sampled from $\pi$, one has $z\in S$ with probability $1$.
For $z\in S$, 
one has $\ln \pi(z) = \ln \pi_0(z)-\ln \pi_0(S)$. Therefore, 
\begin{align}
\sum_{i=1}^g A(z_i)\nabla \ln \pi(z_i)
&=
\sum_{i=1}^g A(z_i)\left(\nabla \ln
\pi_0(z_i)-\nabla \ln \pi_0(S)\right)
\\&=\sum_{i=1}^g A(z_i)\nabla \ln
\pi_0(z_i) -\left(
\sum_{i=1}^g A(z_i)
\right)\nabla \ln \pi_0(S)
\\&=\sum_{i=1}^g A(z_i)\nabla \ln
\pi_0(z_i)
\end{align}
since the sum of advantages in RLOO satisfies $\sum_i A(z_i)=0$.
\end{proof}

\subsection{Reward and Promotion Details}\label{app:training-details}
Algorithm \ref{alg:detailed-appendix-ma} details our full algorithm. Note that MA refers to the rolling moving average.

\begin{algorithm}[htb]
    \small
   \caption{\algoname : Teacher-Student meta-RL Training}
   \label{alg:detailed-appendix-ma}
\begin{algorithmic}
   \STATE {\bfseries Input:} Initial teacher $\pi^T_\phi$, initial student $\pi^S_\theta$, threshold $\tau$, group size $g$, dataset size $n$, repeats $r$, MA window size $m$
   \STATE Initialize timestep $t \gets 0$, MA reward $\bar{R}_0 \gets 0$, $\mathcal{D}_{\text{best}} \gets \emptyset$
   \WHILE{$t < T$}
      \vspace{0.3em}
      \STATE \texttt{\textbf{// 1. Teacher generation}}
      \STATE Sample $g \cdot n$ QA pairs: $\{(q_i, a_i)\}_{i=1}^{g \cdot n} \sim \pi^T_\phi$ 
      \STATE Partition into $g$ datasets: $\mathcal{X}_k = \{(q_{j},a_{j})\}_{j=n(k-1)+1}^{nk}$ for $k=1,\dots,g$
      \STATE Sample reward questions $\mathcal{Q}_R = \{(q_j, a_j)\}_{j=1}^M \sim \mathcal{D}_{\text{train}}$
      
      \vspace{0.3em}
      \STATE \texttt{\textbf{// 2. Inner Loop}}
      \FOR{$k=1$ {\bfseries to} $g$} 
         \FOR{$j=1$ {\bfseries to} $r$}
                \STATE $\theta'_{k,j} \gets \textsc{RLOO-Update}(\theta, \mathcal{X}_k)$ \COMMENT{\texttt{\textbf{Student RL}}}
                \STATE $R_{k,j} \gets \textsc{Acc}(\theta'_{k,j}, \mathcal{Q}_R) - \textsc{Acc}(\theta, \mathcal{Q}_R)$
         \ENDFOR
         
         $R_k \gets \frac{1}{r} \sum_{j=1}^r R_{k,j}$
      \ENDFOR

      \vspace{0.3em}
      \STATE \texttt{\textbf{// 3. Check for student promotion.}}
      \STATE Set $\hat{R}_t \gets \frac{1}{g}\sum_{k=1}^g R_k$
      \STATE Update $\bar{R}_t \gets \textsc{MA}(\hat{R}_{t-m},...,\hat{R}_t)$
      
      \IF{$\bar{R}_{t} > \tau$}
         \STATE $k^* \gets \arg \max_k R_k$
         \STATE Find $j^*$ such that $R_{k^*, j^*}$ is the median reward in $\{R_{k^*, j}\}_{j=1}^r$
         \STATE $\theta \gets \theta'_{k^*, j^*}$ \COMMENT{\texttt{\textbf{Student Promotion}}}
        \STATE $\mathcal{D}_{\text{best}} \gets \mathcal{D}_{\text{best}} \cup \mathcal{X}_{k^*}$
      \ENDIF

      \vspace{0.3em}
      \STATE \texttt{\textbf{// 4. Teacher Policy Update (Outer-loop)}}
      \STATE $\phi \gets \textsc{RLOO-Update}(\phi, \{(\mathcal{X}_k, R_k)\}_{k=1}^g)$ \COMMENT{\texttt{\textbf{Teacher RL}}}
      \STATE $t \gets t + 1$
   \ENDWHILE
   \STATE \textbf{return} $\mathcal{D}_{\text{best}}$, $\pi^S_\theta$
\end{algorithmic}
\end{algorithm}

\textbf{Stabilizing teacher rewards.} Training inner-loop students with RL can potentially lead to noisy trajectories, and thus noisy teacher rewards. To stabilize the teacher rewards, for each sampled dataset $\X_k$ we execute $r$ parallel student trainings and evaluations, and average their rewards to obtain the final reward: $R_k = \frac{1}{r} \sum_{j=1}^r R_{k, j}$. In practice, we use $r=4$.

\textbf{Promotion mechanism.} At each outer-loop timestep we train $r$ students on each dataset $\X_k$, and ``promote" the student baseline when the rolling moving average of aggregated teacher rewards within a fixed window-size exceeds a fixed threshold $\tau$. 
We choose which trained student to promote by selecting the dataset $\X_k$ with the highest reward $R(\X_k)$ and then selecting the student with the median reward amongst those trained on $\X_k$. 

\textbf{Computing student rewards.} For inner-loop and evaluation RL on the student, we use the \textit{Math-Verify} package to compare the student-generated and ground-truth answers \citep{kydlicek2025mathverify}. We assign a reward following standard formulations for RLVR with math:

$R(y, a)$ = 
$\begin{cases} 
120.0 & \text{if } \text{has\_boxed}(y) \land \text{verify}(y, a) \\
20.0  & \text{if } \text{has\_boxed}(y) \land \neg\text{verify}(\dots) \land a \in y_{ans} \\
10.0  & \text{if } \text{has\_boxed}(y) \land \neg\text{verify}(\dots) \land a \notin y_{ans} \\
0.0   & \text{otherwise}
\end{cases}$

\subsection{Learnability Reward.}\label{app:learnability} 
To ablate the effects of our grounded reward versus intrinsic rewards, we train teacher models using the well-studied learnability reward \citep{zhao2025absolute, sukhbaatar2017asymmetric}. We use the same candidate-generation and dataset-partitioning procedure as \algoname. For each candidate dataset $\mathcal{X}_k = \{q_i,a_i)\}_{i=1}^n$, we sample 32 completions from the student for each $q_i$ and compute the average success rate $\bar{s}_i$. The per-question reward is then computed as  
\begin{equation}
r_{i} = \begin{cases} 
0, & \text{if } \bar{s}_{i} = 0 \\
1 - \bar{s}_{i}, & \text{otherwise.}
\end{cases}
\end{equation}

We then compute the dataset-level reward as $R_k = \frac{1}{n} \sum_{i=1}^n r_i$. For consistency with \algoname, every rollout in $\mathcal{X}_k$ receives the averaged dataset-level reward. We train learnability teachers for 200 steps, and observe convergence of rewards. 

\subsection{Datasets}\label{app:datasets}

\textbf{Fail@128 Filtering.} For each problem in the pool of candidates, we sample 128 solutions with \texttt{Llama-3.2-3B-Instruct} using the student prompt in Appendix \ref{app:prompts}, a token budget of 1024 tokens, and temperature 1.0. We keep problems that obtained a 0/128 success rate.

\textbf{OlympiadBench.} For OlympiadBench, we source our fail@128 questions from the subset that is in English, text-only, and automatically verifiable (674 total questions). Since OlympiadBench was originally designed as a test set, we construct a random train/test split. 

\textbf{HARP.} We source our fail@128 problems from the full HARP dataset. Since HARP was originally designed as a test set, we construct a random train/test split.

\textbf{MATH.} In preliminary experiments, we observed a large gap between the zero-shot accuracy of \texttt{Llama-3.2-3B-Instruct} on the official MATH training vs. test splits (60\% vs. 37\%), suggesting that the model may have partial exposure to the MATH training questions. To minimize confounding effects from such memorization, we draw our initial pool of hard problems from the 5000-problem official MATH test split. We then apply the fail@128 filter and construct our own internal train/test split from this filtered subset. All synthetic data generation and student-teacher training uses only the internal training split, and final results are reported exclusively on the held-out internal test split.

\textbf{Dataset sizes.} In Table \ref{tab:app_dataset_sizes} we report the original size of each problem pool, and the sizes of our train/test splits.
\begin{table}[h]
    \centering
    \caption{\centering \textbf{Dataset sizes pre- and post- fail@128 filtering.}}
    \label{tab:app_dataset_sizes}
    \begin{tabular}{l c c c c}
        \toprule
        \textbf{Dataset} & \textbf{Initial problem pool} & \textbf{fail@128 train set} & \textbf{fail@128 test set} \\
        \midrule
        MATH           & 5000 & 359 & 360 \\
        HARP           & 4768 & 714 & 714 \\
        Olympiad Bench & 674 & 158 & 158 \\
        \bottomrule
    \end{tabular}
\end{table}

\subsection{Evaluation}\label{app:eval}

\textbf{Mixed synthetic-real training.} We primarily evaluate generated questions by training a fresh student model on a combination of the synthetic questions, and the real fail@128 train set. We explore two mixing strategies:
\begin{itemize}
    \item \textbf{Curriculum training.} We first train the student on synthetic questions for a fixed number of training steps (64), and then switch to training on real fail@128 training questions, aiming to mirror the trajectory of training a promoted student. Here, the synthetic questions act as a ``warm-start", enabling the student to obtain gradient signal on the harder problems. The synthetic training window was chosen as a representative budget based on preliminary experiments.
    \item \textbf{Mixed training. } We train on a mixture of synthetic and real questions throughout. 
\end{itemize}
To avoid biasing results, we select between curriculum/mixed training using our baseline methods. 

\begin{figure*}[h]
    \centering
\includegraphics[width=0.65\linewidth]{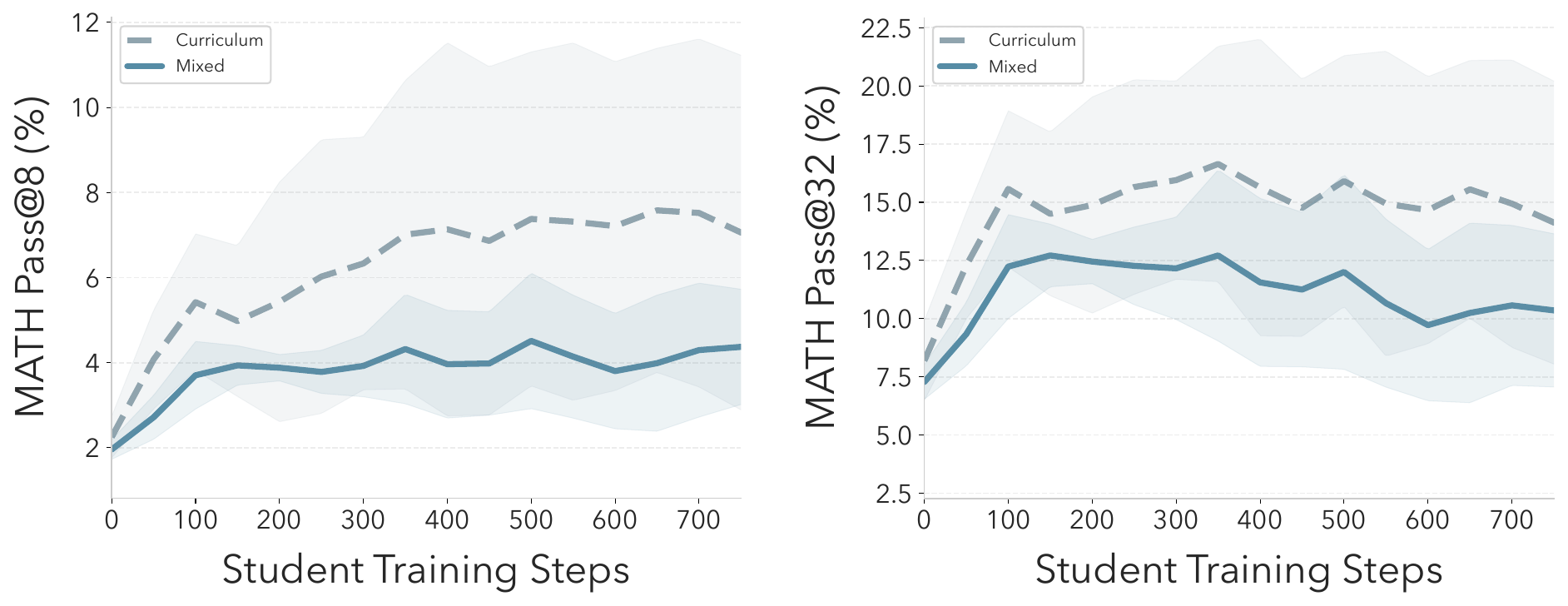}
    \caption{\textbf{Mixed v. Curriculum training on MATH.} We compare training the base student on MATH fail@128 + 128 questions sampled from \base. Curriculum performs better across different inference budgets.}
    \label{fig:math-staged-mixed}
\end{figure*}

\begin{figure*}[h]
    \centering
\includegraphics[width=0.65\linewidth]{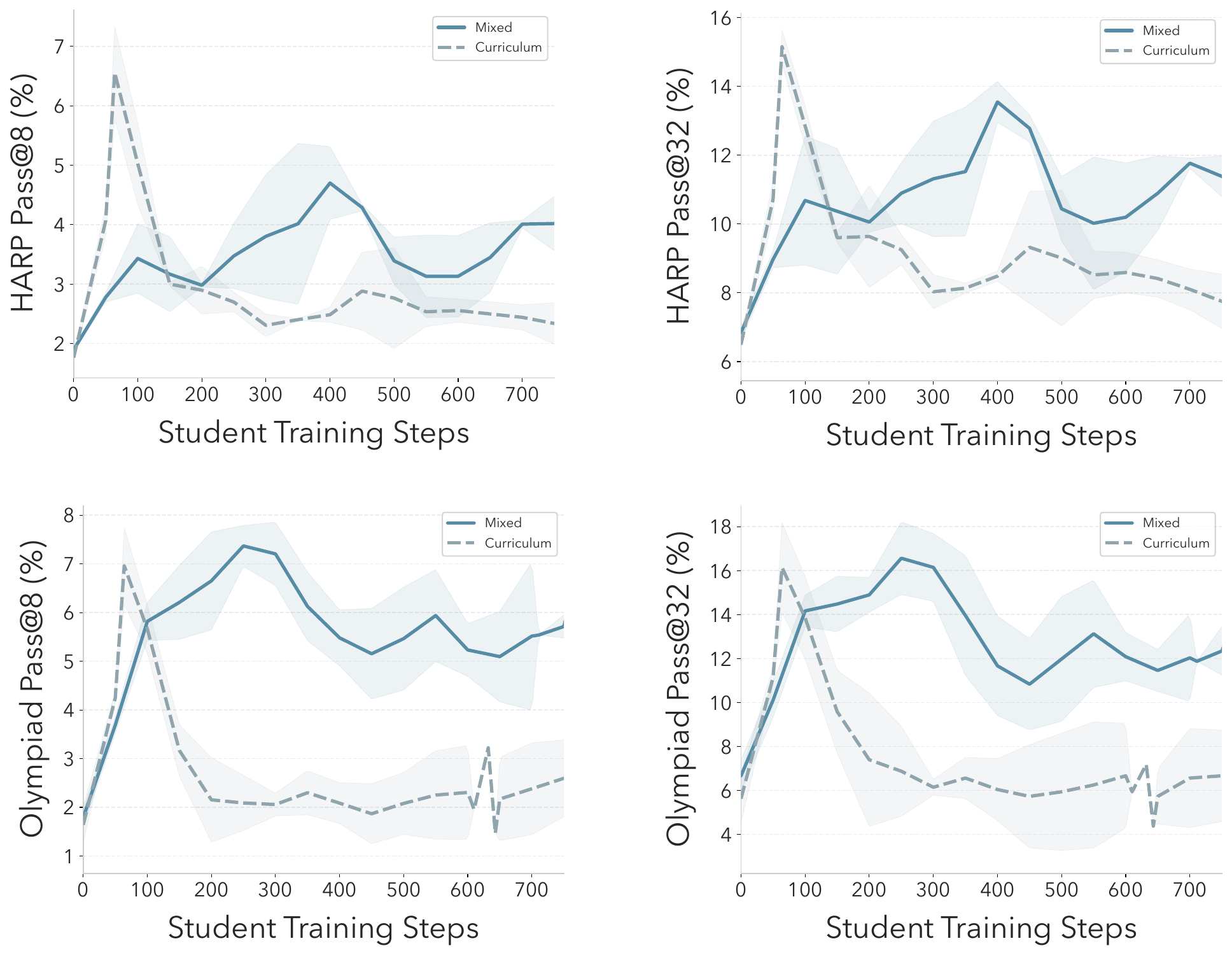}
    \caption{\textbf{Mixed v. Curriculum training on HARP/OlympiadBench.} We compare training the base student on real fail@128 + 128 random MATH questions, for HARP and OlympiadBench. Mixed training exhibits significantly more stable training dynamics across inference budgets (Pass@8 and Pass@32) and converges to higher final performance points. For both datasets, curriculum training exhibits more instability with a large early performance spike and then crash.}
    \label{fig:harp-olympiad-staged-mixed}
\end{figure*}

On MATH, while both exhibit similar training dynamics, we found that our \base baseline performed better with curriculum and thus adopt it for all MATH experiments (Figure \ref{fig:math-staged-mixed}). On OlympiadBench and HARP we observed that mixed training yields significantly more stable learning dynamics, even when adding real instead of synthetic data. 
Figure \ref{fig:harp-olympiad-staged-mixed} compares mixed/curriculum training on HARP and OlympiadBench fail@128 with 128 real MATH problems. Curriculum training exhibits an early performance spike, followed by a significant and sudden performance decline early in training. Thus for HARP and OlympiadBench we use mixed training in our evaluations.

\textbf{Teacher sampling.} At evaluation time, we sample problems from the trained teacher using the same prompt and format-filtering as in training.

\textbf{PQ/PS Evaluation.} We evaluate PQ using mixed synthetic/real training, described above. We evaluate PS by simply running inference on the fail@128 test set, to evaluate how much the student baseline advanced during \algoname training.  

\textbf{Student checkpoint selection.} For evaluations involving fresh student models, we train for a maximum of 1500 steps (typically observing convergence well before this point). 
For MATH and HARP experiments where we report performance at a fixed point, we select the student checkpoint to evaluate at using the \textit{slope of the smoothed training reward curve}, similarly to classic RL early stopping heuristics \citep{Mahsereci2017EarlySW}. In particular, we smooth the average training reward curve (centered-moving-average, 25 steps) and compute the discrete slopes, normalized by the range of observed rewards. The early stopping step is defined as the earliest point where the normalized slope falls below 15\% of the maximum observed slope.
We selected a 15\% threshold to identify the beginning of the reward plateau; empirically, varying between 10\% and 20\% have negligible effects on the selected point.
Test performance is averaged over a 200 step window following the selected step, to account for variance. We also show the full training curves in Section \ref{app:full-training-curves}. 

We choose this heuristic to account for differing convergence rates between methods on MATH and HARP, and our small dataset sizes. In initial experiments we found separate validation sets, and cross-validation with the train set, to be extremely noisy. On OlympiadBench we observe similar convergence across all methods, and report at a fixed point of 50 steps.

\subsection{Hyperparameters} \label{app:hparams}

In Table \ref{tab:hparams} we detail our training and evaluation hyperparameters.

\textbf{Outer-loop training.} We performed the following sweeps in preliminary experiments, and tuned using student performance on the full train set. Once selected, the same hyperparameters are used across all training runs and datasets. See Appendix \ref{app:ab-teacher-training} for ablations on sensitivity to threshold $\tau$ and dataset size $n$.
\begin{itemize}
    \item LR: \{1e-6, 5e-6, \textbf{1e-5}, 5e-5\}
    \item $n$: \{8, 16, 32, \textbf{64}\}
    \item $\tau$: \{\textbf{0.01}, 0.015, 0.02\}
    \item Moving avg window size: \{1, \textbf{3}\}
\end{itemize}

We train for a maximum of 200 outer steps based on compute constraints. For teacher-sampling experiments we fix the evaluation checkpoint based on the point of decline of teacher rewards observed in initial runs (170 steps for all HARP-trained models, 200 steps for all MATH-trained models).

\textbf{Inner-loop training.} We find that from the base student, 10 steps is sufficient to induce movement in student performance. As the student baseline is updated, it is helpful to train slightly longer (we use +5 steps).
We use greedy decoding for evaluating on $\mathcal{Q}_R$ to reduce noise in the student reward.

\textbf{Evaluation.} We use standard hyperparameters to train the student from scratch on combined real/synthetic data (Table \ref{tab:hparams}c). For PQ with curriculum evaluation we use zero learning rate warmup 
to match the inner-loop environment. 

\subsection{Seeds}\label{app:seeds}
To ensure statistical significance and account for both teacher-training and student-training variation, we employ a nested seeding strategy. 

\textbf{Teacher training.} 
\begin{itemize}
    \item For our main \algoname experiments, we train four independent teachers each on MATH and HARP to cover a range of teacher training outcomes.
    \item For teacher objective ablations (\intrinsic and \groundednp) we trained three independent teachers each. 
\end{itemize}

\textbf{Evaluation (student training).}
\begin{itemize}
    \item The \hardonly baseline is evaluated over $\geq 6$ student seeds.
    \item For PQ datasets ($>$2 promotions), we train at least three students per PQ dataset, totaling $\ge 6$ seeds (2 PQ datasets $\times$ 3 students) per reported metric. 
    \item For PS students, we compute pass@$k$ metrics using inference over three seeds. 
    \item For teacher-sampling experiments (\textit{i.e.,} sampling data from trained teachers and then training a fresh student) we train 2-3 independent students per teacher seed, resulting in $\geq 8$ seeds per reported metric.
\end{itemize}

For all metrics we report the aggregated mean and standard deviation over student seeds.

\begin{table}[htbp]
    \centering
    \caption{\textbf{Hyperparameters for \algoname training and evaluation.}}
    \small
    \begin{tabular}{l c c}
        \toprule
        \textbf{Hyperparameter} & \textbf{Teacher} & \textbf{Student} \\
        \midrule
        Optimizer & \multicolumn{2}{c}{AdamW} \\
        KL coefficient & \multicolumn{2}{c}{0.001} \\
        LR schedule & \multicolumn{2}{c}{Cosine decay} \\
        Learning rate & \multicolumn{2}{c}{1e-5} \\
        Temperature & \multicolumn{2}{c}{1.0} \\
        LR warmup steps & 20 & 0/20 \\
        Batch size & 2 & 8 \\
        Group size & 4 & 32 \\
        Max generated tokens & 512 & 1024 \\
        \midrule
        \multicolumn{3}{l}{\textit{meta-RL specific (teacher only)}} \\
        Promotion threshold ($\tau$) & 0.01 & — \\
        Moving avg window & 3 & — \\
        Dataset size ($n$) & 64 & — \\
        Student repeats ($r$) & 4 & — \\
        \midrule
        \multicolumn{3}{l}{\textit{Evaluation specific (student only)}} \\
        Max training steps & — & 1500 \\
        Synthetic warmup steps & — & 64 \\
        (curriculum training) &  &  \\
        \bottomrule
    \end{tabular}
    \label{tab:hparams}
\end{table}

\subsection{Computational Resources} \label{app:compute}
Each \algoname \textit{training} run was executed on 4 nodes (each 8$\times$ NVIDIA H200 GPUs or 8$\times$ NVIDIA H100 GPUs) for $\approx$ 48-60 hours. Each RLOO \textit{evaluation} run (training a fresh student) was executed for $\approx$ 12 hours on 1 H200 node or 1 H100 node.

\begin{figure*}[h!]
    \centering
\includegraphics[width=\linewidth]{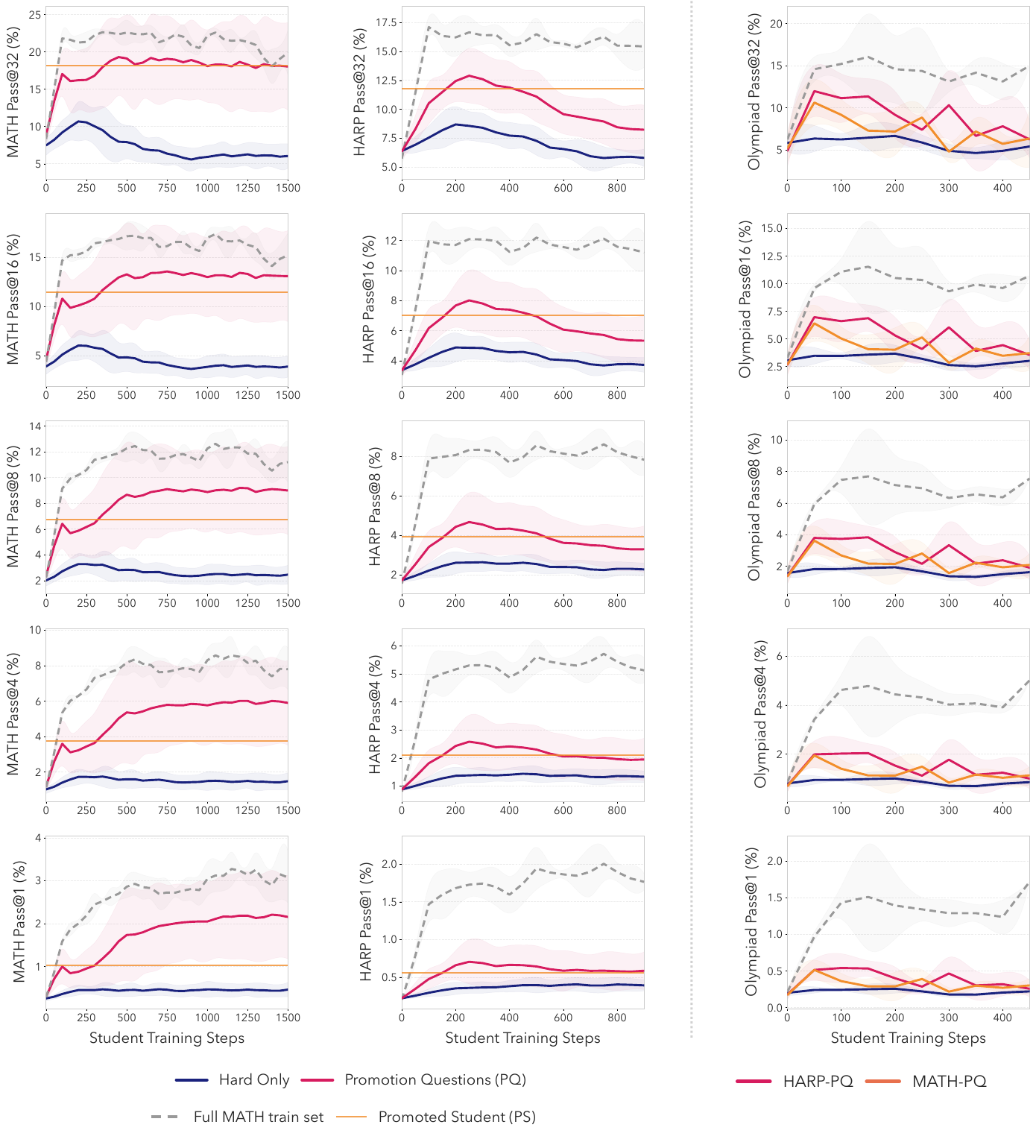}
    \caption{\textbf{Fail@128 test performance during student training for MATH, HARP, and Olympiad.} Student learning curves for different pass@k when trained on \hardonly, PQ, or the Full MATH dataset (PS inference performance shown as a horizontal line). PQ and PS improve performance on all inference budgets and datasets, with increased effect at higher $k$. On MATH, PQ exhibits performance gains even after the synthetic-training phase (64 steps), showing that synthetic problems make real hard problems more learnable. }
    \label{fig:accepted-qs-full}
\end{figure*}

\begin{figure*}[h]
    \centering
\includegraphics[width=1.0\linewidth]{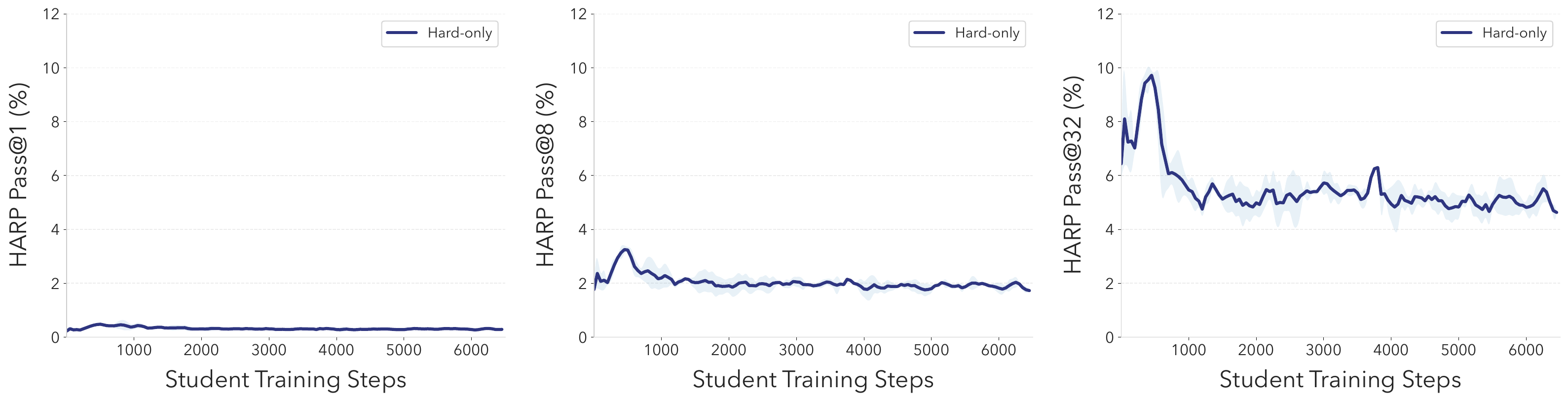}
    \caption{\textbf{Extended \hardonly training on HARP.} We extend \hardonly training on HARP from 1500 steps to 6500 steps, and find that the extra compute does not improve performance. Shading shows $\pm$ 1 SD over 3 seeds.}
    \label{fig:harp-hardonly-extended}
\end{figure*}

\begin{figure*}[h]
    \centering
\includegraphics[width=0.8\linewidth]{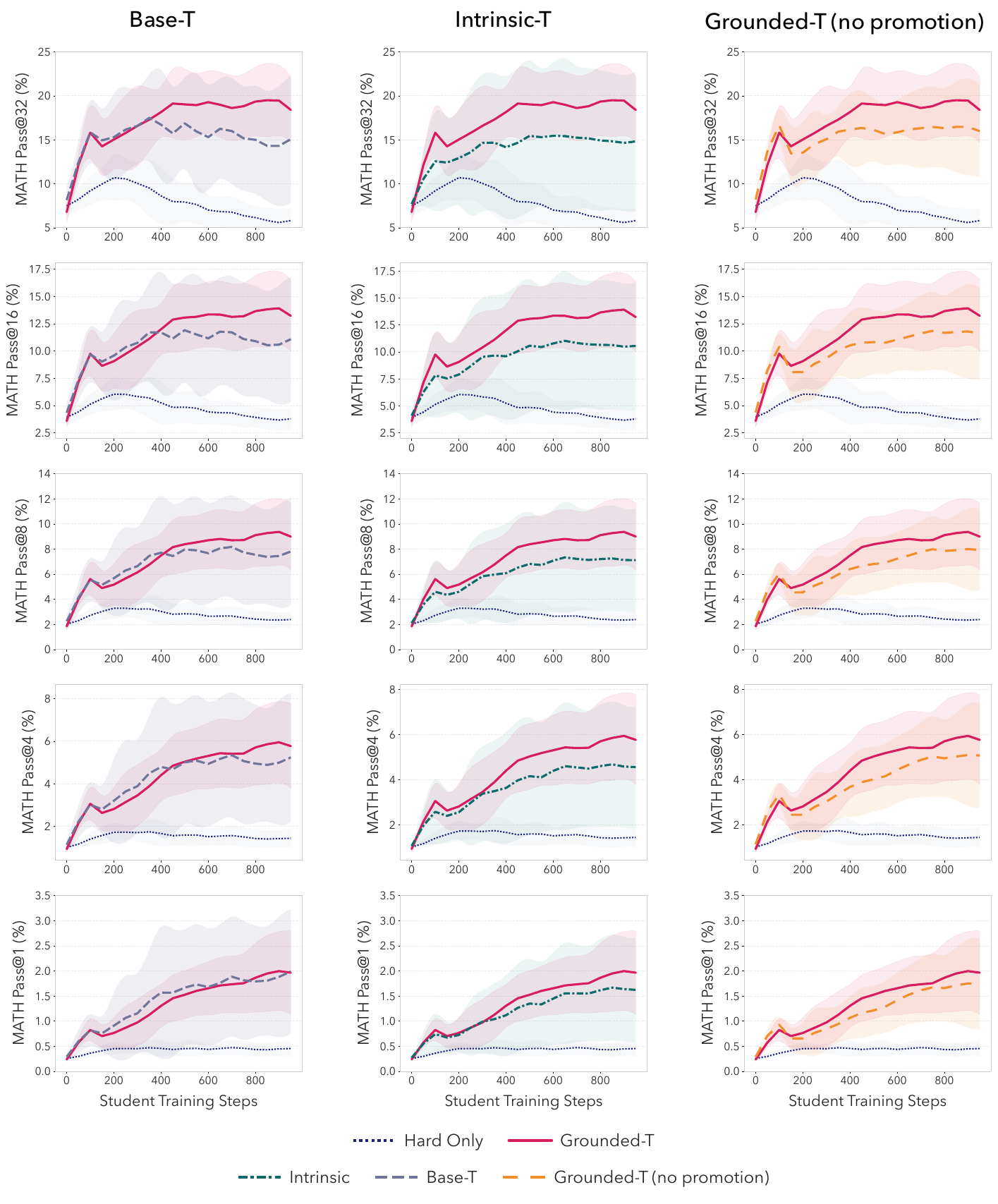}
    \caption{\textbf{Fail@128 test performance during student training for MATH with different teachers.} Each column compares training a fresh student with 128 questions from \grounded to 128 questions from a different teacher (\hardonly also included for reference). While all teachers outperform \hardonly, \grounded performs best, with increasing effects at higher $k$. \grounded results in less variance across student outcomes, particularly compared to \base and \intrinsic. PQ learning curves are in Figure \ref{fig:accepted-qs-full}. }
    \label{fig:teacher-ablation-MATH}
\end{figure*}

\begin{figure*}[h]
    \centering
\includegraphics[width=0.8\linewidth]{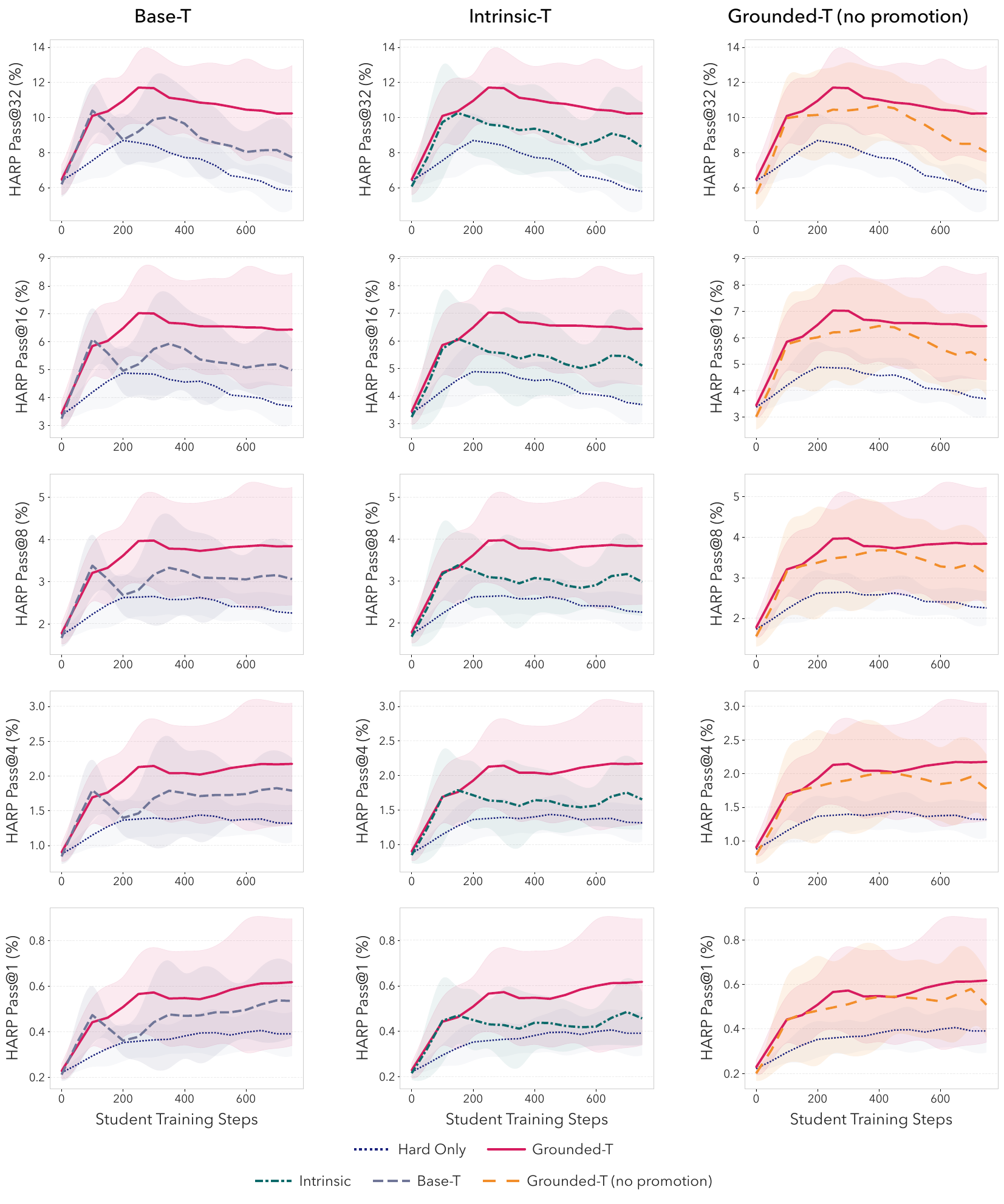}
        \caption{\textbf{Fail@128 test performance during student training for HARP with different teachers.} Each column compares training a fresh student with 128 questions from \grounded to 128 questions from a different teacher (\hardonly also included for reference). \grounded performs best, with increasing effects at higher $k$. Students trained with \base and \intrinsic tend to decline more for higher $k$ in the later stages of training, while \grounded leads to more stable trajectories.}
    \label{fig:teacher-ablation-HARP}
\end{figure*}

\begin{figure*}[h]
    \centering
\includegraphics[width=0.7\linewidth]{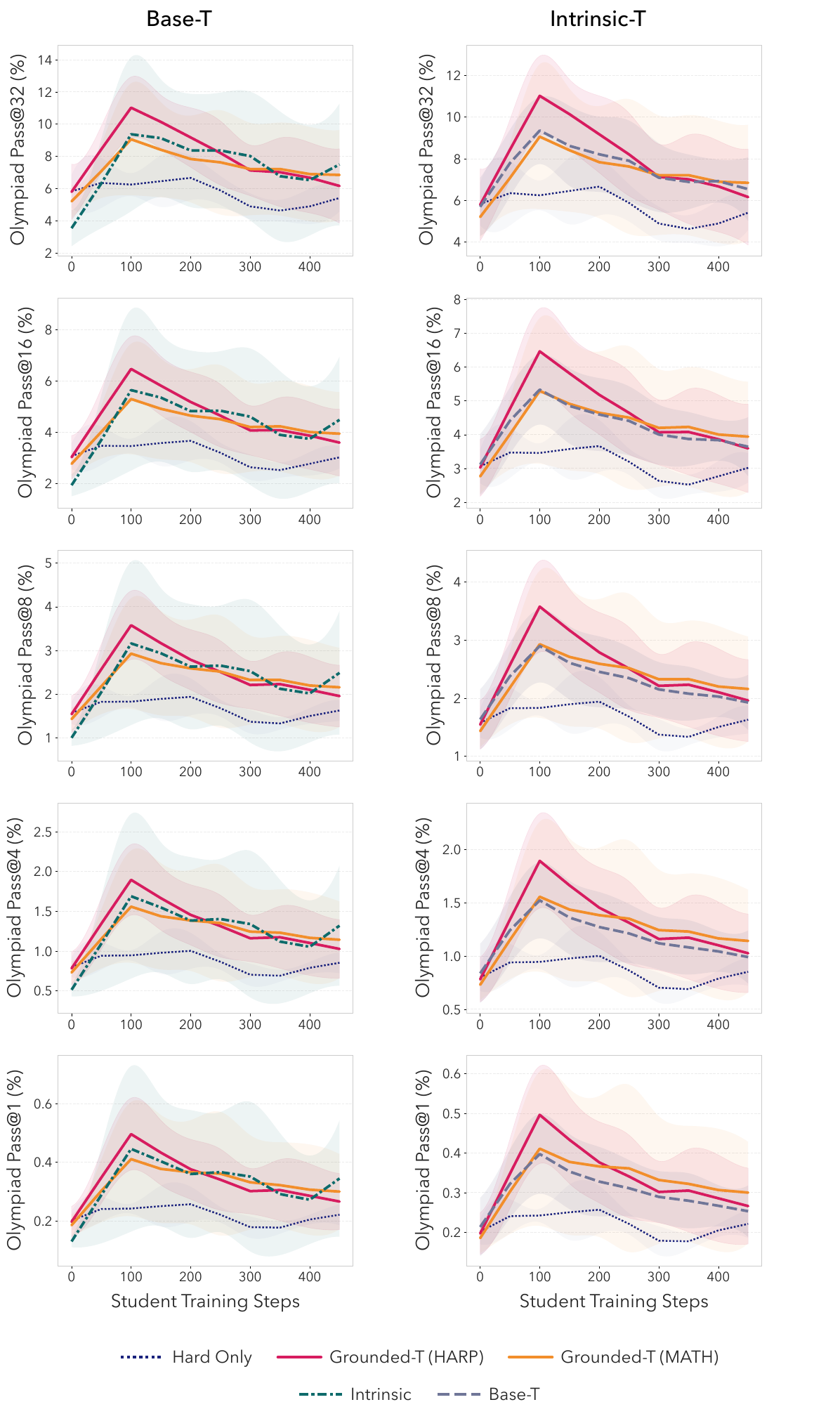}
    \caption{\textbf{Fail@128 test performance during student training for Olympiad with different teachers.} Each column compares training a fresh student with 128 questions from \grounded (trained with MATH and HARP) to 128 questions from a different teacher (\hardonly also included for reference). Students trained with \grounded teachers have more similar mean performance to \base and \intrinsic than seen on HARP and MATH (Figures \ref{fig:teacher-ablation-MATH}-\ref{fig:teacher-ablation-HARP}). However, \textit{\grounded (HARP) shows more stability and less variance between independent teachers than \intrinsic (see Figure \ref{fig:teacher-ablation-ckpt-olympiad})}.}
    \label{fig:teacher-ablation-olympiad}
\end{figure*}

\begin{figure*}[h]
    \centering
\includegraphics[width=0.8\linewidth]{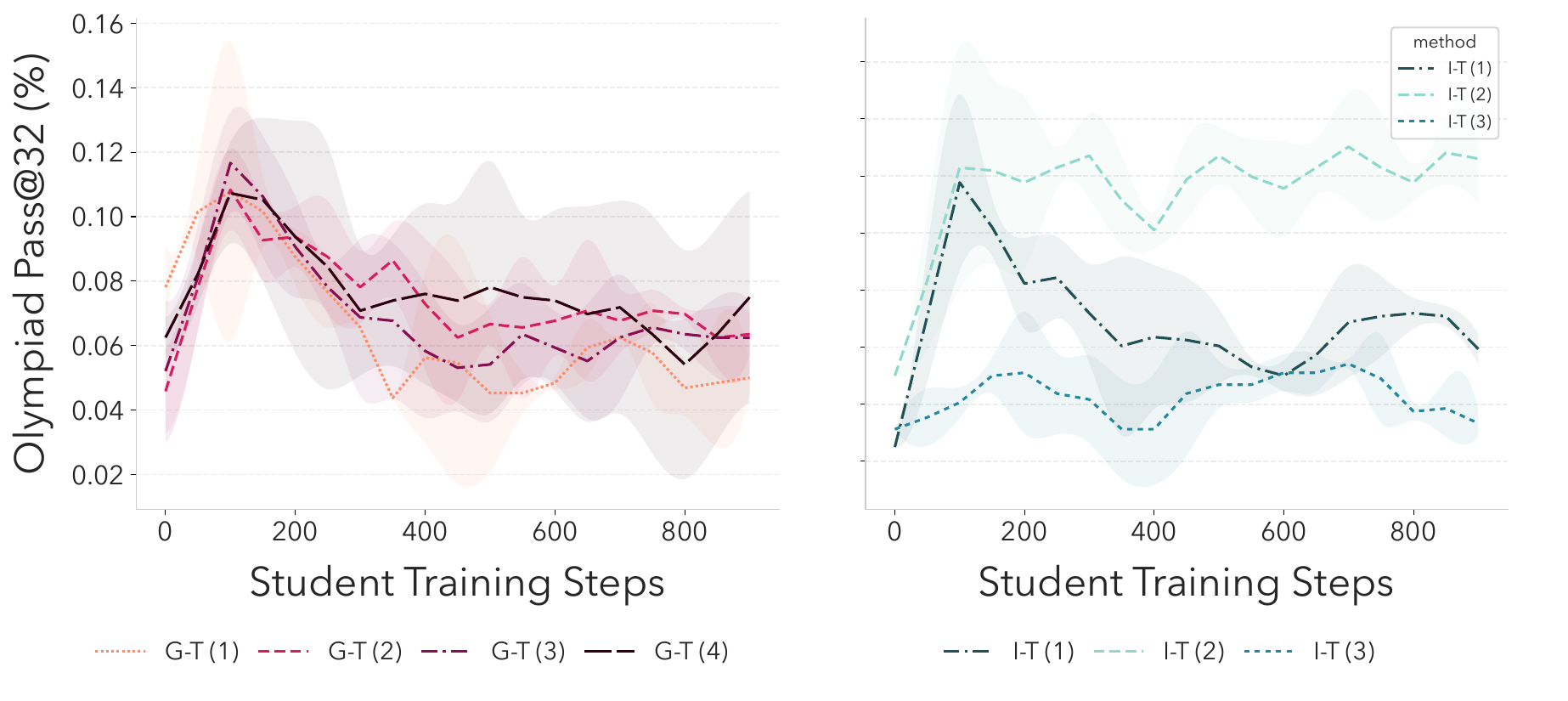}
    \caption{\textbf{Test Pass@32 on OlympiadBench for fresh students trained with individual \grounded teacher seeds (red) and \intrinsic teacher seeds (green).} Questions from \grounded yield consistent student trajectories on OlympiadBench across different teachers, whereas \intrinsic exhibits high variance across teachers, including a failure mode where I-T (1) causes student collapse. }
    \label{fig:teacher-ablation-ckpt-olympiad}
\end{figure*}

\section{Expanded Main Experiments}

\subsection{Full Student Training Curves}\label{app:full-training-curves}
In Figure \ref{fig:accepted-qs-full} we show full student training curves for PQ, \hardonly, and the full MATH upper bound for MATH, HARP, and OlympiadBench. In Figures \ref{fig:teacher-ablation-MATH}-\ref{fig:teacher-ablation-olympiad} we show these training curves for questions sampled from \grounded, \base, \intrinsic, and \groundednp. All curves show the mean and standard deviation over seeds. \textbf{}

\subsection{Full Evaluations on fail@128 MATH, HARP, and OlympiadBench.}\label{app:pq-full}
\begin{table}[h]
\centering
\caption{\textbf{MATH pass@k (\%) test accuracy on fail@128.} Mean and SD over seeds are reported at the timestep determined by training reward convergence (see Appendix \ref{app:eval}) with full curves in Figure \ref{fig:accepted-qs-full}. PQ and PS consistently outperform  inference-only, \hardonly, and intrinsic baselines across all inference budgets, and recover the majority of performance gain from training with real curated problems. 
We boldface the best among ``data-free" methods (\textit{i.e., only $\mathcal{D}_{train}$ available}). The bottom three rows serve as upper bounds from using curated, expert-annotated data. PQ datasets contain one of $\{128, 192, 256\}$ questions.}
\label{tab:app-promotion-results-math}
\begin{tabular}{l ccccc}
\toprule
& \multicolumn{5}{c}{\textbf{k}} \\
\cmidrule(lr){2-6}
\textbf{Method} & \textbf{1} & \textbf{4} & \textbf{8} & \textbf{16} & \textbf{32} \\
\midrule
Base Model Inference
& $0.3 \pm 0.1$ & $1.0 \pm 0.2$ & $2.0 \pm 0.4$ & $3.9 \pm 0.8$ & $7.5 \pm 1.3$ \\
\midrule
\hardonly
& $0.5 \pm 0.1$ & $1.7 \pm 0.4$ & $3.2 \pm 0.8$ & $5.7 \pm 1.5$ & $9.6 \pm 2.6$ \\
\hardonly ($g=128$)
& $1.4 \pm 1.0$ & $3.9 \pm 2.6$ & $6.1 \pm 3.9$ & $8.9 \pm 5.5$ & $12.4 \pm 7.4$ \\
\midrule
SeRL (step 50)
& $0.5 \pm 0.1$ & $1.9 \pm 0.3$ & $3.7 \pm 0.4$ & $6.9 \pm 0.7$ & $12.2 \pm 0.7$ \\
SeRL (step 100)
& $0.5 \pm 0.0$ & $2.1 \pm 0.0$ & $4.0 \pm 0.0$ & $7.3 \pm 0.2$ & $12.4 \pm 1.0$ \\
\midrule
\algoname-PQ (Ours)
& $\mathbf{1.7 \pm 1.0}$ & $\mathbf{5.3 \pm 2.6}$ & $\mathbf{8.5 \pm 3.7}$ & $13.0 \pm 4.8$ & $18.9 \pm 5.3$ \\
\algoname-PS (Ours)
& $1.0 \pm 0.2$ & $3.8 \pm 0.6$ & $6.8 \pm 1.1$ & $11.5 \pm 1.6$ & $18.1 \pm 2.4$ \\
\grounded (Ours)
& $1.6 \pm 0.5$ & $5.1 \pm 1.4$ & $8.4 \pm 2.1$ & $\mathbf{13.1 \pm 2.9}$ & $\mathbf{19.1 \pm 3.7}$ \\
\intrinsic
& $1.0 \pm 0.6$ & $3.3 \pm 2.1$ & $5.7 \pm 3.5$ & $9.2 \pm 5.3$ & $14.1 \pm 7.5$ \\
Inference with \grounded
& $0.3 \pm 0.0$ & $1.1 \pm 0.2$ & $2.2 \pm 0.3$ & $4.4 \pm 0.6$ & $8.3 \pm 1.1$ \\
\midrule
\midrule
HARP train (128)
& $2.4 \pm 1.0$ & $7.2 \pm 2.4$ & $11.3 \pm 3.1$ & $16.5 \pm 3.6$ & $23.0 \pm 3.9$ \\
MATH train (128)
& $2.1 \pm 0.0$ & $6.6 \pm 0.1$ & $10.5 \pm 0.3$ & $15.7 \pm 0.5$ & $21.8 \pm 0.9$ \\
MATH train (Full)
& $2.7 \pm 0.2$ & $7.6 \pm 0.7$ & $11.5 \pm 1.2$ & $16.4 \pm 1.8$ & $22.0 \pm 2.4$ \\
\bottomrule
\end{tabular}
\end{table}

\begin{table}[ht]
\centering
\caption{\textbf{HARP pass@k (\%) test accuracy on fail@128}. Mean and SD over seeds are reported at the timestep determined by training reward convergence (see Appendix \ref{app:eval}) with full curves in Figure \ref{fig:accepted-qs-full}. PQ and PS consistently outperform  inference-only, \hardonly, and intrinsic baselines across all inference budgets. Notably, \algoname questions perform better on HARP than similar numbers of questions from the MATH/HARP datasets (which serve as a curated, expert-annotated data source).}
\label{tab:app-promotion-results-harp}
\begin{tabular}{l ccccc}
\toprule
& \multicolumn{5}{c}{\textbf{k}} \\
\cmidrule(lr){2-6}
\textbf{Method} & \textbf{1} & \textbf{4} & \textbf{8} & \textbf{16} & \textbf{32} \\
\midrule
Base Model Inference
& $0.2 \pm 0.0$ & $0.9 \pm 0.0$ & $1.7 \pm 0.0$ & $3.4 \pm 0.0$ & $6.4 \pm 0.0$ \\
\midrule
\hardonly
& $0.4 \pm 0.1$ & $1.4 \pm 0.2$ & $2.6 \pm 0.4$ & $4.7 \pm 0.6$ & $8.2 \pm 1.0$ \\
\midrule
SeRL (step 50)
& $0.4 \pm 0.0$ & $1.4 \pm 0.1$ & $2.7 \pm 0.3$ & $5.3 \pm 0.5$ & $9.7 \pm 0.9$ \\
SeRL (step 100)
& $0.5 \pm 0.0$ & $1.9 \pm 0.1$ & $3.6 \pm 0.2$ & $6.5 \pm 0.3$ & $10.9 \pm 0.4$ \\
\midrule
\algoname-PQ (Ours)
& $\mathbf{0.7 \pm 0.3}$ & $\mathbf{2.5 \pm 0.8}$ & $\mathbf{4.5 \pm 1.3}$ & $\mathbf{7.7 \pm 1.7}$ & $\mathbf{12.3 \pm 2.0}$ \\
\algoname-PS (Ours)
& $0.6 \pm 0.1$ & $2.1 \pm 0.3$ & $3.9 \pm 0.6$ & $7.0 \pm 0.9$ & $11.8 \pm 1.2$ \\
\grounded (Ours)
& $0.5 \pm 0.2$ & $2.0 \pm 0.5$ & $3.8 \pm 0.9$ & $6.7 \pm 1.3$ & $11.2 \pm 1.7$ \\
\intrinsic
& $0.4 \pm 0.1$ & $1.6 \pm 0.5$ & $3.1 \pm 0.8$ & $5.6 \pm 1.4$ & $9.6 \pm 2.1$ \\
Inference with \grounded
& $0.2 \pm 0.0$ & $0.9 \pm 0.1$ & $1.9 \pm 0.2$ & $3.6 \pm 0.4$ & $6.8 \pm 0.7$ \\
\midrule
\midrule
HARP train (128)
& $0.4 \pm 0.0$ & $1.4 \pm 0.1$ & $2.8 \pm 0.2$ & $5.0 \pm 0.5$ & $8.7 \pm 1.1$ \\
MATH train (128)
& $0.6 \pm 0.1$ & $2.1 \pm 0.4$ & $4.0 \pm 0.7$ & $7.1 \pm 0.9$ & $11.9 \pm 0.9$ \\
MATH train (Full)
& $1.7 \pm 0.2$ & $5.1 \pm 0.4$ & $8.1 \pm 0.4$ & $11.7 \pm 0.3$ & $16.2 \pm 0.4$ \\
\bottomrule
\end{tabular}
\end{table}

\begin{table}[ht]
\centering
\caption{\textbf{Olympiad pass@k (\%) test accuracy on fail@128.} Mean and SD over seeds are reported timestep 50 with full curves in Figure \ref{fig:accepted-qs-full}. Despite being optimized with reward signals from HARP and MATH, PQ questions and PS inference transfer to improving performance on Olympiad, and match or outperform 128 questions sampled from the HARP train set (a curated/expert-annotated source of problems). PS and PQ transfer better when trained with HARP than with MATH, potentially indicating more shared structure between HARP and Olympiad.}\label{tab:app-promotion-results-olympiad}
\begin{tabular}{l ccccc}
\toprule
& \multicolumn{5}{c}{\textbf{k}} \\
\cmidrule(lr){2-6}
\textbf{Method} & \textbf{1} & \textbf{4} & \textbf{8} & \textbf{16} & \textbf{32} \\
\midrule
Base Model Inference
& $0.2 \pm 0.0$ & $0.8 \pm 0.1$ & $1.6 \pm 0.3$ & $3.1 \pm 0.5$ & $5.8 \pm 1.0$ \\
\hardonly
& $0.3 \pm 0.1$ & $1.1 \pm 0.3$ & $2.1 \pm 0.6$ & $3.9 \pm 1.3$ & $6.9 \pm 2.7$ \\
\midrule
\algoname-PQ (MATH) (Ours)
& $0.5 \pm 0.1$ & $1.9 \pm 0.5$ & $3.6 \pm 0.9$ & $6.4 \pm 1.6$ & $10.6 \pm 2.7$ \\
\algoname-PQ (HARP) (Ours)
& $0.5 \pm 0.1$ & $2.0 \pm 0.5$ & $\mathbf{3.8 \pm 1.0}$ & $\mathbf{7.0 \pm 1.8}$ & $\mathbf{12.0 \pm 3.0}$ \\
\algoname-PS (MATH) (Ours)
& $\mathbf{0.6 \pm 0.1}$ & $\mathbf{2.1 \pm 0.5}$ & $3.7 \pm 0.8$ & $6.2 \pm 1.3$ & $9.9 \pm 2.2$ \\
\algoname-PS (HARP) (Ours)
& $0.5 \pm 0.1$ & $2.0 \pm 0.4$ & $\mathbf{3.8 \pm 0.7}$ & $6.9 \pm 1.1$ & $11.7 \pm 1.6$ \\
\grounded (MATH) (Ours)
& $0.4 \pm 0.2$ & $1.6 \pm 0.8$ & $2.9 \pm 1.4$ & $5.3 \pm 2.4$ & $9.0 \pm 4.0$ \\
\grounded (HARP) (Ours)
& $0.5 \pm 0.2$ & $1.9 \pm 0.6$ & $3.6 \pm 1.1$ & $6.5 \pm 1.8$ & $11.1 \pm 2.9$ \\
\intrinsic
& $0.4 \pm 0.3$ & $1.7 \pm 1.2$ & $3.1 \pm 2.0$ & $5.5 \pm 3.4$ & $9.1 \pm 5.2$ \\
\midrule
\midrule
HARP train (128)
& $0.5 \pm 0.1$ & $2.0 \pm 0.2$ & $3.6 \pm 0.4$ & $6.5 \pm 0.8$ & $10.6 \pm 1.7$ \\
MATH train (128)
& $1.0 \pm 0.1$ & $3.4 \pm 0.1$ & $5.9 \pm 0.1$ & $9.6 \pm 0.4$ & $14.6 \pm 1.4$ \\
MATH train (Full)
& $0.9 \pm 0.0$ & $3.2 \pm 0.1$ & $5.6 \pm 0.3$ & $8.8 \pm 0.7$ & $13.1 \pm 0.9$ \\
\bottomrule
\end{tabular}
\end{table}

In Tables \ref{tab:app-promotion-results-math}-\ref{tab:app-promotion-results-harp} we report our full results from evaluating \algoname on MATH and HARP (in-domain datasets). In Table \ref{tab:app-promotion-results-olympiad} we report full results from evaluating on OlympiadBench, an OOD dataset. 

Our PQ datasets have one of $\{128, 192, 256\}$ questions, depending on the number of student promotions for each run. For \intrinsic we sample 128 questions, consistent with all of our teacher-sampling experiments. For the equal-data comparison between \intrinsic and \grounded (sampling from the \algoname-trained teacher), see Section \ref{sec:teacher-ablation} and Appendix \ref{app:teacher-ablation}.

In addition to the methods/baselines shown in Figure \ref{fig:promotion-results}  we also report the following.

\textbf{Inference pass@k with the base model.} Inference with the base model has non-zero pass@$k$ due to stochastic sampling with different seeds than were used for the initial pass$@128=0$ filtering. Comparison with \hardonly results shows that our fail@128 datasets are sufficiently difficult such that direct training yields very little improvement. 

\textbf{Inference pass@k with the trained teacher.} We perform direct inference on fail@128 problems with final trained teacher policies, denoted as ``Inference with \grounded" in Tables \ref{tab:app-promotion-results-math}-\ref{tab:app-promotion-results-harp}). We find no improvement over the base model, indicating that improved \textit{teaching ability} is decoupled from improved \textit{solving ability}.

\textbf{\hardonly with extra compute.} A natural question is whether we can improve direct training on fail@128 train questions simply by increasing compute. One strategy is to train for longer, however our learning curves in Figure \ref{fig:accepted-qs-full} show that \hardonly test performance decreases in the latter stages of training. We confirm this by extending \hardonly training with HARP from 1500 to 6500 steps (Figure \ref{fig:harp-hardonly-extended}). Another strategy is to sample more from the base model by increasing the RLOO group size. On MATH, we increase the group size $4\times$ (from our default $g=32$ to $g=128$), and find that it only yields marginal improvements over \hardonly (\textit{e.g., } +2.8\% pass@32) and does not recover the improvements of PQ. 

\textbf{SeRL.} We evaluate SeRL \cite{fang2025serl} as a contemporary self-play baseline. SeRL self-evolves a curriculum using an initial seed dataset with the goal of outperforming direct RLVR on that dataset. SeRL generates question-answer pairs with majority-voting self-rewards, and filters based on learnability and diversity. We train SeRL with MATH and HARP fail@128 respectively as the seed sets, and observe convergence after 100 steps. As \texttt{Llama-3.2-3B-Instruct} is a primary model used in \cite{fang2025serl}, we adopt their reported hyperparameters. Results are shown in Tables \ref{tab:app-promotion-results-math} and \ref{tab:app-promotion-results-harp}; SeRL improves over \hardonly however fails to match \algoname.

\textbf{Sampling curated ``oracle questions".} In addition to training with the full MATH train set, we also evaluate sampling 128 questions from the MATH and HARP train sets, which can be considered oracle (curated/expert-annotated) data sources. We choose 128 to match our teacher sampling experiments (Section \ref{app:teacher-ablation}) and roughly match the amount of PQ data, which varies between 128 and 256 questions. 

On MATH, training with these smaller subsets performs similarly to training with the full MATH dataset, suggesting a saturation point. On HARP, these smaller subsets only recover $\approx 50\%$ of the gains from training with the full MATH train set. Notably, PQ and PS both outperform 128 sampled questions from HARP, and match 128 questions from MATH.

\subsection{Sampling from Teacher Models.}\label{app:teacher-ablation}
While PQ comes from accumulated useful questions over the meta-RL trajectory, here we \textit{sample questions directly from the trained teacher policy}. The similar performance of \grounded and PQ (Tables \ref{tab:app-promotion-results-math}-\ref{tab:app-promotion-results-harp}) provide evidence that the pedagogical signals captured in the PQ datasets are learned by the teacher's distribution. 

In Figures \ref{fig:teacher-ablation-MATH}-\ref{fig:teacher-ablation-olympiad} we show full test trajectories on MATH, HARP, and Olympiad for students trained with 128 questions sampled from \grounded, \intrinsic, \base, and \groundednp. \grounded outperforms all comparisons, particularly at higher inference budgets, and is competitive with PQ. \grounded also exhibits lower variance and greater stability across student and teacher seeds. \groundednp performs worse than \grounded, PQ, and PS, validating the importance of the promotion mechanism. 

In Figure \ref{fig:teacher-ablation-ckpt-olympiad} we also compare student trajectories for each \grounded and \intrinsic teacher seed. Consistent with MATH and HARP (Figure \ref{fig:teacher-ablation}), students have similar trajectories across independent \grounded teachers, and high variance across different \intrinsic teachers, showcasing the instability of intrinsic rewards. 

\subsection{Categorizing Correctness of Synthetic Questions.}\label{app:taxonomy}

We categorize synthetic questions into \textit{correctness taxonomies} using \texttt{Claude-4.5-Sonnet} as an oracle judge. The prompt given to Claude is shown below. In Table \ref{tab:app-error-analysis-final} we report taxonomy statistics for PQ datasets, and problems sampled from \grounded, \intrinsic, and \base teachers.  

We prompt \texttt{Claude-4.5-Sonnet} to categorize problems as follows:
\begin{itemize}
    \item \texttt{Well posed}: If the problem is mathematically complete and solvable.
    \item \texttt{Correct}: If the proposed answer is correct (only if the problem is well posed).
    \item \texttt{Error type}:
    \begin{itemize}
        \item \texttt{None}
        \item \texttt{Arithmetic error}: Sound logic, but incorrect final calculation.
        \item \texttt{Logical fallacy}: Does not follow mathematical rules.
        \item \texttt{Ill-posed/Impossibility}: The question contains a mathematical impossibility.
        \item \texttt{Ambiguous}: The question is missing data, variables, or context necessary for solving it.
    \end{itemize}
\end{itemize}

Our results show that the well-posedness of a problem matters more than the correctness of the solution. While teacher-training improves the correctness rate, the best-performing datasets (\grounded and PQ) only contain 32.8\% and 36.5\% correct solutions respectively, compared to 55.5\% for \intrinsic. This indicates that question diversity is more important for success (see Table \ref{tab:final_diversity_results}). Meta-RL mainly reduces question ambiguities (improving well-posedness) while the rate of arithmetic errors remains the same or slightly higher. 

\subsection{Do Incorrect Synthetic Questions Help?}\label{app:correctness-ablation}
We run a controlled experiment to isolate the effects of training on well-posed questions with incorrect answers. We compare training a fresh student on the subset of PQ-HARP questions with correct answers (\textit{Correct-only}; 82 questions), and the subset with well-posed questions (\textit{Well-posed}; 144 questions). Note that \textit{Correct-only} is strictly a subset of \textit{Well-posed}. In both cases, we follow the standard HARP evaluation protocol and train on a mix of synthetic and real fail@128 questions. 

\begin{figure*}[h]
    \centering
\includegraphics[width=0.65\linewidth]{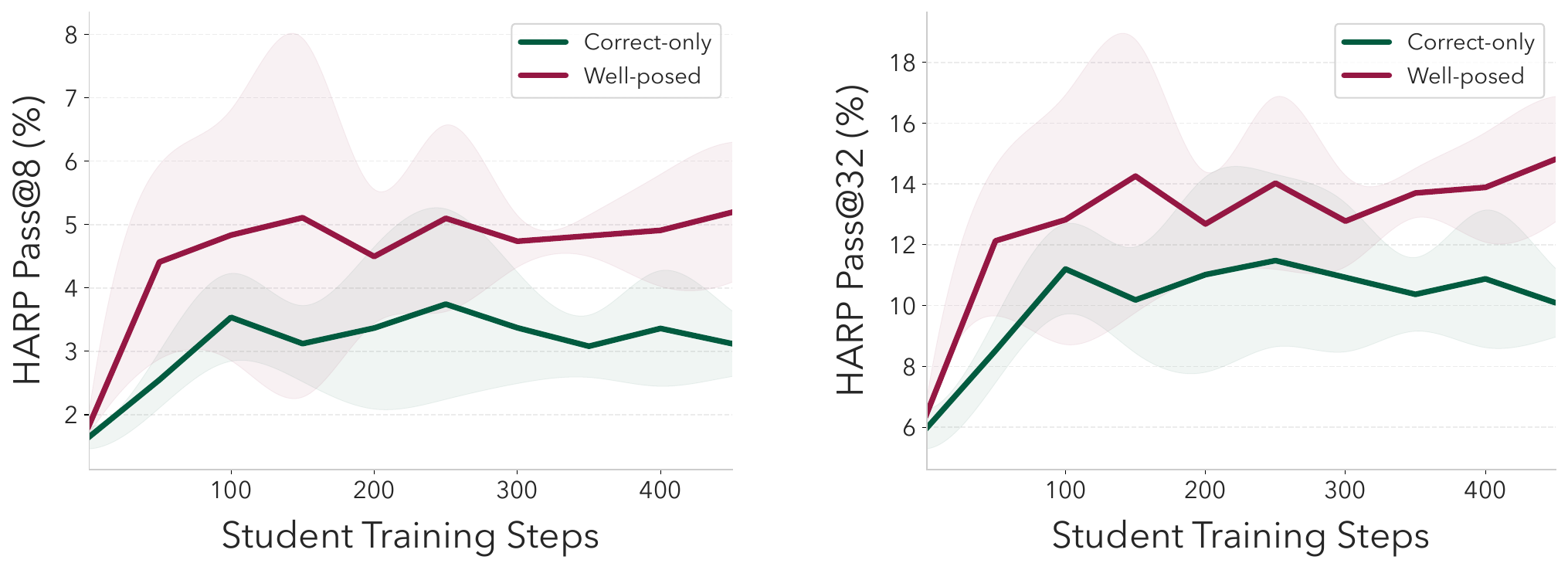}
    \caption{\textbf{Training on Correct-only v. Well-posed questions (HARP).} We compare training the base student on real fail@128 + \textit{Correct-only} HARP-PQ questions, and real fail@128 + \textit{Well-posed} HARP-PQ questions. Performance improves when adding well-posed questions with incorrect answers. Shading shows $\pm$ 1 SD over 3 seeds.}
    \label{fig:harp-correctness-ablation}
\end{figure*}

Results are shown in in Figure \ref{fig:harp-correctness-ablation}. Adding well-posed questions with incorrect answers improves performance, showing that performance gains are not driven solely by correct questions. We further note that across all experiments (\textit{Hard-Only}, \textit{Correct-only}, \textit{Well-posed}) the student format match reaches 100\% within the first $\sim 50$ steps; thus, formatting is also not the main driver of performance gains. 

\begin{tcblisting}{
  arc=0mm,
  outer arc=0mm,
  colback=gray!5,
  colframe=gray!50,
  listing only, 
  listing options={style=tcblatex, breaklines=true, basicstyle=\ttfamily\small},
  title=Oracle Prompt
}
You are evaluating generated math problems for their coherence and solvability. Your task is to determine if the given question is well-formulated, and if the given answer is correct.

CRITICAL INSTRUCTION: Do not assume missing information. If the question is nonsensical, lacks a clear problem/question/equation, is syntactically incorrect, is missing necessary information, or is missing variables, you MUST classify it as 'Ambiguous' or 'Ill_Posed'. Do not invent a context to make the answer work.

QUESTION: {question}
PROPOSED_ANSWER: {proposed_answer}

TAXONOMY OF ERRORS:
- 'None': The question is mathematically complete and the answer is correct.
- 'Arithmetic': The logic is sound, but the final calculation is wrong.
- 'Logical_Fallacy': The steps taken do not follow mathematical rules.
- 'Ill_Posed': The question contains a mathematical impossibility.
- 'Ambiguous': The question is missing necessary data, variables, or context (e.g., "Solve the equation" without providing the equation).

TASK:
1. Analyze the QUESTION for completeness. If it's a "fragment" or "nonsense," stop and flag it.
2. Solve the problem ONLY if it is well-defined.
3. Determine:
   - is_well_posed: boolean - Is the question mathematically complete and solvable?
   - is_correct: boolean - Is the proposed answer correct? (Only evaluate if is_well_posed is true)
   - error_type: one of ['None', 'Arithmetic', 'Logical_Fallacy', 'Ill_Posed', 'Ambiguous']
   - verified_answer: string - The correct answer if the question is well-posed, or "N/A" if not well-posed

OUTPUT FORMAT:
First, provide your reasoning in <think> tags.
Then, provide a JSON object with the following exact structure:

```json
{{
    "is_correct": <boolean>,
    "is_well_posed": <boolean>,
    "error_type": "<one of: None, Arithmetic, Logical_Fallacy, Ill_Posed, Ambiguous>",
    "verified_answer": "<string: the correct answer or 'N/A'>"
}}
``

EXAMPLE OUTPUT:
<think>
The question asks to solve 2x + 5 = 13. This is well-posed with all necessary information. Solving: 2x = 8, so x = 4. The proposed answer is 4, which is correct.
</think>

```json
{{
    "is_correct": true,
    "is_well_posed": true,
    "error_type": "None",
    "verified_answer": "4"
}}
\end{tcblisting}
\clearpage

\begin{table}[ht]
\centering
\caption{\textbf{Correctness analysis and error taxonomy of synthetic questions, evaluated by \texttt{Claude-4.5-Sonnet}}. Teacher training (for both grounded and intrinsic rewards) improves the well-posedness and correctness of problems relative to the base model, with a corresponding decrease in question ambiguity errors. \grounded and PQ have fewer correct questions than \intrinsic but perform better, potentially because of greater diversity (see Table \ref{tab:final_diversity_results}.)}\label{tab:app-error-analysis-final}
\begin{tabular}{lcccc}
\toprule
\textbf{Category} & \textbf{Base} & \textbf{Intrinsic} & \textbf{Grounded} & \textbf{PQ} \\
\midrule
Well-Posed & 53.6\% & 63.5\% & 70.0\% & 64.6\% \\
Correct    & 23.2\% & 55.5\% & 36.5\% & 32.8\% \\
\midrule
\multicolumn{2}{l}{\textbf{Error Taxonomy (\% of total samples)}} & & & \\
\midrule
Arithmetic Error & 23.7\% & 5.7\%  & 29.0\% & 25.0\% \\
Logic Error      & 5.7\%  & 2.3\%  & 6.9\%  & 6.5\%  \\
Impossibility Error  & 4.7\%  & 2.9\%  & 8.2\%  & 4.7\%  \\
Ambiguity Error  & 42.4\% & 33.6\% & 21.3\% & 31.3\% \\
\midrule
Total Samples    & 384    & 384    & 375    & 384    \\
\bottomrule
\end{tabular}
\end{table}

\section{Ablations}

\subsection{Sampled dataset size} \label{app:sampled-dataset-size}
\begin{figure*}[ht]
    \centering
\includegraphics[width=0.8\linewidth]{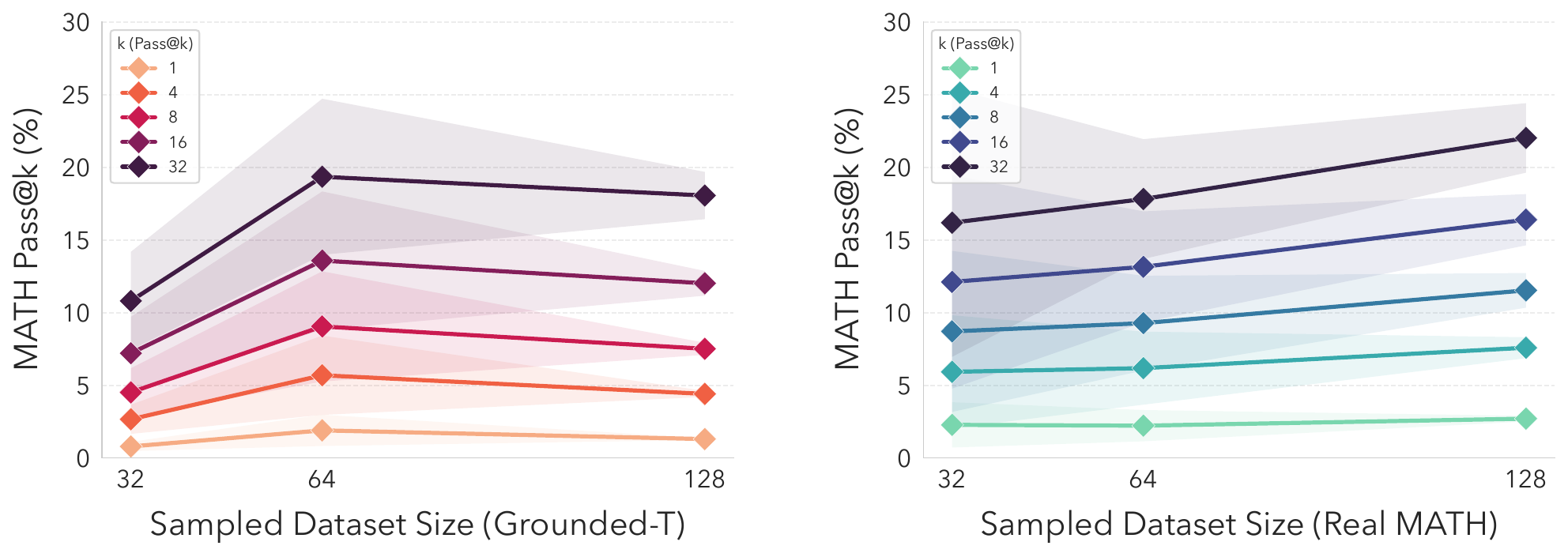}
    \caption{\textbf{(Left) Sampling different-sized datasets from \grounded for MATH (fail@128).} Mean and $\pm$ 1 SD across 2 teacher seeds and 2 student seeds. \textbf{(Right) Sampling different-sized datasets from the full MATH train set for MATH (fail@128).} Resampled for each seed, 3 seeds.}\label{fig:n-ablation}
\end{figure*}

When training with \algoname, teacher-generated problems are partitioned into datasets that the student is trained on in the inner loop. Thus the teacher rewards are based on a specific dataset size (64 in our case). In evaluation, however, one could potentially sample any number of questions from the teacher policy. This raises the question of how the performance of sampled datasets changes with size. Is it best to sample the number of questions that the teacher was trained with, or does performance saturate at higher sampling rates? 

We evaluate two teacher models trained with MATH by sampling $n \in \{32, 64, 128\}$ questions from each teacher, and training a fresh student on the sampled questions and the MATH fail@128 train set (3 seeds per run). Since teacher models are trained with $n=64$, this covers datasets smaller, equal to, and larger than the dataset size that the teacher was trained with. 

Results are shown in Figure \ref{fig:n-ablation} for different pass@$k$. Performance improves with increasing $n$. Sampling with 128 questions has a similar \textit{mean} performance as sampling 64 questions but with significantly smaller \textit{error}. This illustrates benefits (namely, consistency/reliabilty) to sampling questions from the teacher at higher rates than it was trained with. As a comparison we also perform the same experiment using \textit{real} questions from the MATH training dataset. For all values of $n$, real MATH questions perform similarly or better, and exhibit diminishing variance with increasing numbers of questions. 

\subsection{Sensitivity to Teacher Hyperparameters} \label{app:ab-teacher-training}

We ablate $\tau$ (the teacher-reward threshold to determine if the student baseline should be promoted) and $n$ (the number of samples per dataset that teacher-generated problems are partitioned into). The teacher generates $g \cdot n$ problems per outer-RLOO iteration.

\begin{figure*}[h]
    \centering
\includegraphics[width=1.0\linewidth]{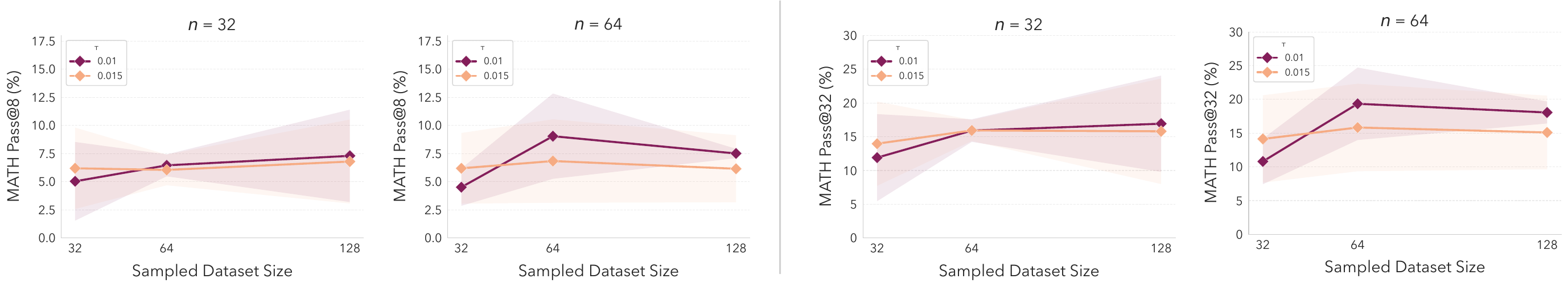}
    \caption{\textbf{Hyperparameter sensitivity on MATH.} We train \algoname with $\tau \in \{0.01, 0.015\}$ and $n \in \{32, 64\}$, then evaluate by training students on datasets of size $|\mathcal{X}| \in \{32, 64, 128\}$. Shaded regions indicate $\pm$1 SD.}
    \label{fig:teacher-hparams}
\end{figure*}

We train \algoname on MATH with $\tau \in \{0.01, 0.015\}$ and $n \in \{32, 64\}$. For each combination we train two \algoname runs for 200 steps and evaluate the final teacher checkpoints by sampling varying amounts of questions ($|\mathcal{X}| \in \{32, 64, 128\}$) and training two fresh students. Results are shown in Figure \ref{fig:teacher-hparams} for pass@8 and pass@32. 
Our default configuration ($n$=64, $\tau$=0.01) performs best, with $n=64$ showing modest
advantages over $n=32$ at larger evaluation dataset 
sizes, which is consistent with the teacher being trained to produce larger datasets.

\subsection{Problem Generation Format.} 
\begin{table}[ht]
\centering
\caption{\textbf{MATH pass@$k$ (\%) test accuracy on fail@128 for multi-turn teacher sampling.} We report mean and SD across four teacher seeds and 2 student seeds per teacher. Multiturn performs worse than our default single-turn setting across all pass@k and sampled dataset sizes. }
\label{tab:qa_results}
\small
\begin{tabular}{ll ccccc}
\toprule
& & \multicolumn{5}{c}{\textbf{k}} \\
\cmidrule(lr){3-7}
\textbf{$n$} & \textbf{$|\mathcal{X}|$} & \textbf{1} & \textbf{4} & \textbf{8} & \textbf{16} & \textbf{32} \\
\midrule
32 & 32  & $\mathbf{0.7 \pm 0.6}$ & $\mathbf{2.3 \pm 1.9}$ & $\mathbf{4.2 \pm 3.1}$ & $\mathbf{7.1 \pm 4.8}$ & $\mathbf{11.4 \pm 6.7}$ \\
   & 64  & $0.5 \pm 0.3$ & $1.9 \pm 0.9$ & $3.6 \pm 1.6$ & $6.4 \pm 2.7$ & $11.0 \pm 4.0$ \\
   & 128 & $0.7 \pm 0.7$ & $2.3 \pm 2.0$ & $4.0 \pm 3.3$ & $6.8 \pm 4.9$ & $11.1 \pm 7.1$ \\
\midrule
64 & 32  & $0.4 \pm 0.1$ & $1.6 \pm 0.4$ & $3.0 \pm 0.8$ & $5.2 \pm 1.4$ & $8.6 \pm 2.5$ \\
   & 64  & $0.4 \pm 0.0$ & $1.5 \pm 0.2$ & $2.9 \pm 0.3$ & $5.3 \pm 0.5$ & $9.4 \pm 0.8$ \\
   & 128 & $0.4 \pm 0.1$ & $1.6 \pm 0.4$ & $2.8 \pm 0.6$ & $4.8 \pm 0.9$ & $8.0 \pm 1.3$ \\
\bottomrule
\end{tabular}
\end{table}

In our default setup, we sample problems from the teacher by prompting it to produce a single completion that is parsed into a question/answer, and filtering out outputs that do not match the necessary format. An alternative sampling method, however, is to generate problems in separate question-answer stages (multi-turn) such that filtering is not needed:
\begin{enumerate}
    \item Sample $\pi_{\phi}^T(q_i |p)$ where $p$ is a teacher prompt to generate a question.
    \item Sample $\pi_{\phi}^T(a_i |p, q_i, p')$ where $p'$ is a prompt to generate an answer given the question.
\end{enumerate}

The logprob component of the teacher RLOO loss is then $\log(\pi_{\phi}^T(q_i|p)) + \log(\pi_{\phi}^T(a_i|p, q_i, p'))$.

We execute \algoname across four seeds using this teacher-sampling formulation with our standard procedure and hyperparameters, ablating $n \in \{32, 64\}$. We observe that the teacher reward quickly plateaus and does not exceed one promotion. In Table \ref{tab:qa_results} we find that across different numbers of sampled problems and values of $n$, the multi-turn sampling strategy performs worse than our default single-turn sampling. 

\subsection{Model Size} \label{app:model-size}
\begin{figure*}[h]
    \centering
\includegraphics[width=1.0\linewidth]{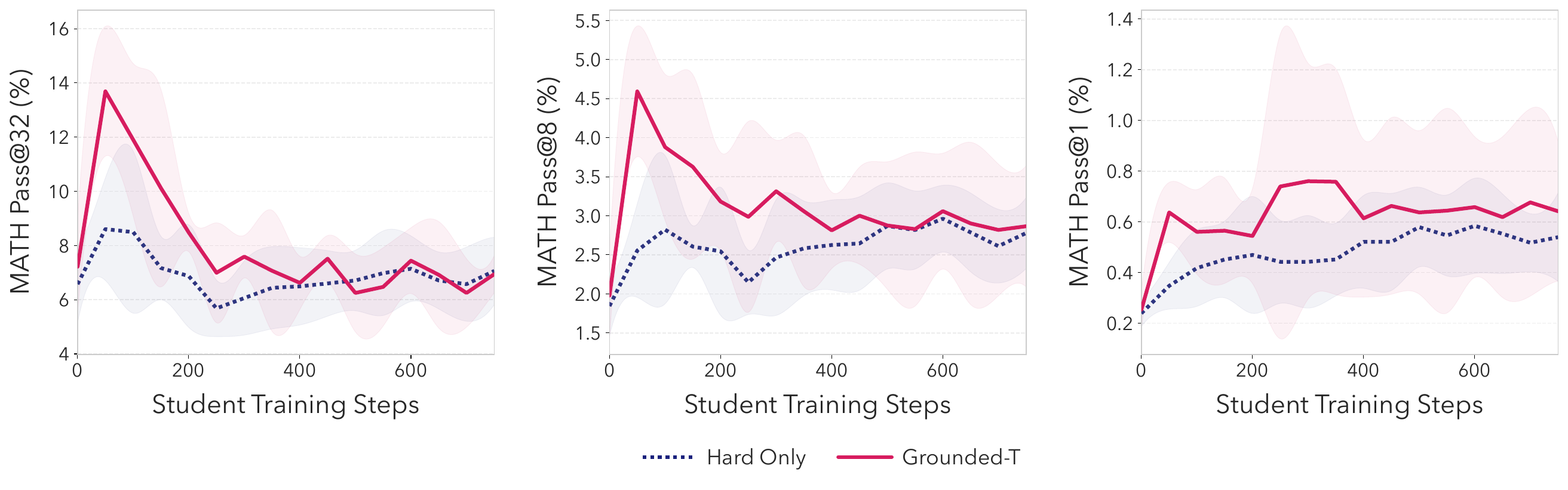}
    \caption{\textbf{Fail@128 test performance during student training for MATH with Llama-3.1-8B-Instruct.} We compare training a fresh student with 128 questions from \grounded to \hardonly. \grounded outperforms \hardonly, showing that key trends transfer to larger models. Shading shows $\pm$ 1 SD across 4 seeds (2 teacher seeds, 2 student seeds per teacher).}
    \label{fig:8b-curves}
\end{figure*}

We train \algoname with \texttt{Llama3.1-8B-Instruct} on MATH fail@128 with the same hyperparameters as those used for \texttt{Llama3.2-3B-Instruct}, due to the compute cost of hyperparameter tuning for meta-RL runs. With these default hyperparameters we do not observe promotions, however find that performance still improves over the \hardonly baseline, indicating that the key trend transfers to bigger models. Results are shown in Table \ref{tab:grounded-t-8b-results}, with student training curves in Figure \ref{fig:8b-curves}.

\begin{table}[ht]
\centering
\caption{\textbf{MATH pass@k (\%) test Accuracy on fail@128 with \texttt{Llama-3.1-8B-Instruct}}. Mean and SD are shown over 2 teacher seeds and 2 student seeds per teacher, at the timestep determined by training reward convergence. \grounded outperforms \hardonly across all inference budgets.}
\label{tab:grounded-t-8b-results}
\begin{tabular}{l ccccc}
\toprule
& \multicolumn{5}{c}{\textbf{k}} \\
\cmidrule(lr){2-6}
\textbf{Method} & \textbf{1} & \textbf{4} & \textbf{8} & \textbf{16} & \textbf{32} \\
\midrule
Base Model Inference
& $0.3 \pm 0.1$ & $1.0 \pm 0.2$ & $2.0 \pm 0.4$ & $3.9 \pm 0.7$ & $7.3 \pm 1.1$ \\
\hardonly
& $0.4 \pm 0.1$ & $1.5 \pm 0.8$ & $2.7 \pm 1.0$ & $5.0 \pm 1.8$ & $9.1 \pm 3.2$ \\
\grounded (Ours)
& $\mathbf{0.6 \pm 0.1}$ & $\mathbf{2.4 \pm 0.5}$ & $\mathbf{4.6 \pm 1.0}$ & $\mathbf{8.2 \pm 1.7}$ & $\mathbf{13.7 \pm 2.7}$ \\
\bottomrule
\end{tabular}

\end{table}

\section{Teacher Training Dynamics}\label{app:teacher-training-curves}
\begin{figure*}[h]
    \centering
\includegraphics[width=0.7\linewidth]{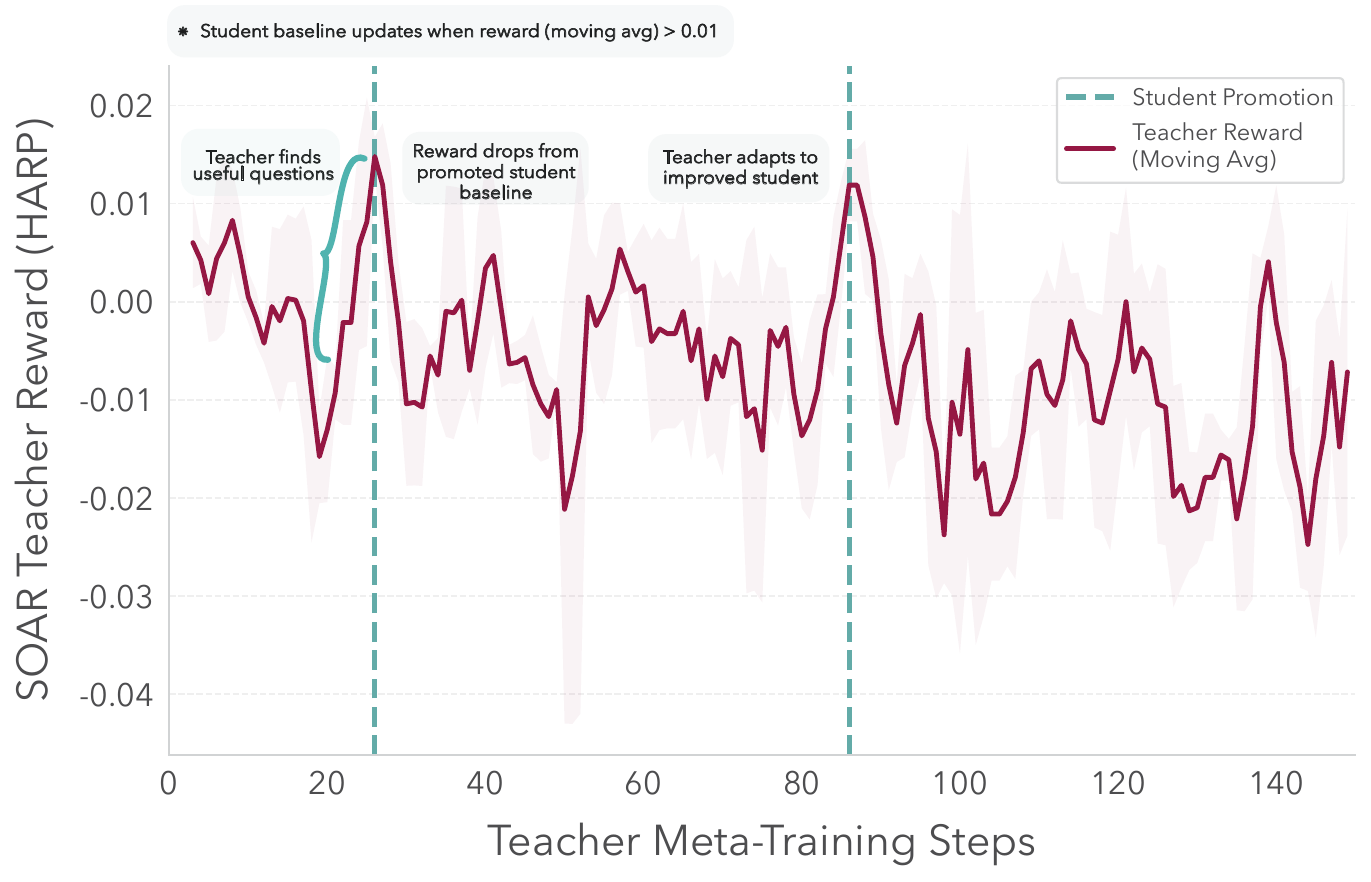}
    \caption{\textbf{Annotated teacher reward dynamics when training \algoname with HARP.} Shows a sample teacher trajectory from a \algoname run on HARP. The teacher follows a cyclical search-exploitation pattern. Student promotions (updating the student baseline to a trained student) are triggered when the 3-step moving average of teacher rewards exceeds $\tau=0.01$. After each promotion, the improved student baseline makes previous curricula less useful, causing rewards to drop, and then recover as the teacher adapts and discovers questions appropriate for the improved student. }
    \label{fig:teacher-rewards}
\end{figure*}

In Figure \ref{fig:teacher-rewards} we show a representative teacher training curve for \algoname on HARP. We observe that \algoname follows a pattern of search and exploitation. The training curve exhibits periods of oscillation (search), and then a steady rise in reward from steps 18-27, culminating in a student promotion. The reward declines after the promotion, due to the improved student baseline, oscillates as the teacher adapts to the improved student, and then exhibits another rise from steps 80-86 culminating in a second promotion.

\begin{figure*}[h]
    \centering
\includegraphics[width=1.0\linewidth]{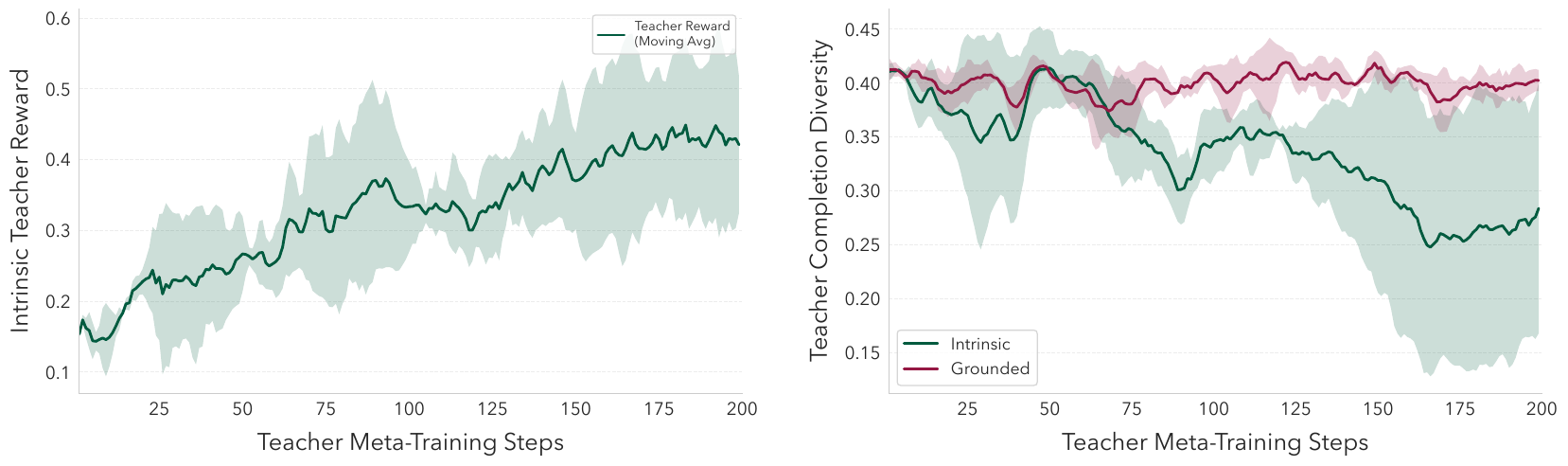}
    \caption{\textbf{(Left) Teacher training dynamics when training with \intrinsic.} Mean and $\pm$ 1 SD over three independent training runs. \textbf{(Right) Teacher completion diversity when training with intrinsic v. grounded rewards.} Grounded rewards preserve diversity for the full run, while intrinsic teachers lose diversity as they converge. Mean and $\pm$ 1 SD over three training runs for intrinsic and four for grounded (two MATH, two HARP).}
    \label{fig:teacher-diversity}
\end{figure*}

Figure \ref{fig:teacher-diversity}a shows teacher training curves for \intrinsic teachers, aggregated across teacher seeds, which exhibits a smooth upward climb. Figure \ref{fig:teacher-diversity}b shows that as the \intrinsic reward climbs, the diversity of teacher completions falls (diversity measured as the average pairwise cosine distance of embeddings). Meanwhile \grounded preserves the original model diversity throughout the full trajectory. This is consistent with findings in Section \ref{sec:teacher-ablation} (Table \ref{tab:final_diversity_results}) that \grounded achieves similar question diversity to \base, whereas \intrinsic teachers collapse to a more narrow conceptual space.

\end{document}